%% file: main.tex
\documentclass[dvipsnames]{article}

\PassOptionsToPackage{numbers,sort&compress,square}{natbib}
\usepackage[final]{lychee}
\setcitestyle{numbers,square,comma,sort&compress}
\usepackage[utf8]{inputenc} % allow utf-8 input
\usepackage[T1]{fontenc}    % use 8-bit T1 fonts
\usepackage{url}            % simple URL typesetting
\usepackage{booktabs}       % professional-quality tables
\usepackage{amsfonts}       % blackboard math symbols
\usepackage{nicefrac}       % compact symbols for 1/2, etc.
\usepackage{multirow}
\usepackage{wrapfig}
\usepackage{amssymb}
\usepackage{epigraph}
\usepackage{makecell}
\usepackage{listings}
\usepackage{fvextra}
\usepackage{subcaption}
\usepackage{algorithm}
\usepackage{algorithmic}
\usepackage[labelfont=bf]{caption} % For bold caption labels

\usepackage{color, soul}
\usepackage{microtype}      % microtypography
\usepackage{xcolor}         % colors
\usepackage{geometry}
\usepackage{mathpazo}
\usepackage[xcharter,bigdelims,vvarbb,noamssymbols,nosymbolsc]{newtxmath}
\usepackage[scaled=1.1]{zlmtt}
\usepackage{ragged2e}
\usepackage{amsmath}
\usepackage{bm}
\usepackage{latexsym}
\usepackage{graphicx}
\usepackage{float}
\usepackage{longtable}
\usepackage{tabularx}
\usepackage{tabu}
\usepackage{bbding}
\usepackage{awesomebox}
\usepackage[most,breakable]{tcolorbox}
\usepackage{etoolbox}
\usepackage{fancyhdr}
\usepackage{lipsum}  
\usepackage{CJKutf8}
\usepackage{pdfpages}
\usepackage{multicol}
\usepackage{pgffor} 
\usepackage{fontawesome5}
\usepackage{nicematrix} 
\usepackage{paralist}
\usepackage[edges]{forest}
\usepackage{siunitx}
\usepackage{tikz-dependency}
\usepackage{tikz}
\usepackage{bbm}
\usepackage[english]{babel}
\usepackage{hyphenat}

\newcommand{\Rmnum}[1]{\expandafter\@slowromancap\romannumeral #1@}

\numberwithin{equation}{section}
\usepackage{chngcntr}
\usepackage[section]{placeins}
\usepackage{afterpage}

\usepackage[colorlinks, linkcolor=RoyalBlue, anchorcolor=BrickRed, citecolor=RoyalBlue, urlcolor=RoyalBlue]{hyperref}

\usepackage{cleveref}
\crefname{section}{\S}{\S\S}
\Crefname{section}{\S}{\S\S}

\newcommand\refsec[1]{Section~\hyperref[sec:#1]{\ref{sec:#1}}}
\newcommand\refsecs[2]{\hyperref[sec:#1]{\S\ref{sec:#1}:~\textsc{#1}}, \hyperref[sec:#2]{\S\ref{sec:#2}:~\textsc{#2}}}

\usepackage[tikz]{bclogo}
\definecolor{msftBlue}{RGB}{0,164,239}
\definecolor{msftGreen}{RGB}{127,186,0}
\definecolor{msftYello}{RGB}{255,185,0}
\definecolor{mypurple}{RGB}{138,43,226} 
\definecolor{msftBlack}{RGB}{0,0,0}

\newtcolorbox{myboxnote}[1][]{
  breakable,
  title=#1,
  colback=cyan!0,
  colbacktitle=cyan!0,
  coltitle=black,
  fonttitle=\bfseries,
  bottomrule=0pt,
  toprule=0pt,
  leftrule=1.5pt,
  rightrule=1.5pt,
  titlerule=0pt,
  arc=0pt,
  outer arc=0pt,
  colframe=lightgray,
}
\usepackage{mdframed}

\definecolor{academicblue}{RGB}{54, 95, 145}

\tcbset{
  iclrtakeawaybox/.style={
    width=\linewidth,
    colback=gray!5,
    colframe=gray!60,
    colbacktitle=gray!30,
    coltitle=black,
    fonttitle=\bfseries,
    boxrule=0.5pt,
    arc=2pt,
    sharp corners=south,
    attach boxed title to top left={yshift=-0.1in,xshift=0.15in},
    boxed title style={boxrule=0pt, colframe=white},
    enhanced,
    drop shadow southeast
  }
}
\newtcolorbox{TakeawayBox}[2][]{iclrtakeawaybox,title=#2,#1}

\definecolor{promptframegray}{RGB}{124,124,124}
\definecolor{promptbodygray}{RGB}{249,249,249}
\newtcolorbox{PromptBox}[2][]{
  enhanced jigsaw,
  breakable,
  width=\linewidth,
  colback=promptbodygray,
  colframe=promptframegray,
  colbacktitle=promptframegray,
  coltitle=white,
  fonttitle=\fontfamily{phv}\selectfont\bfseries,
  fontupper=\small,
  title={#2},
  boxrule=0.7pt,
  arc=3pt,
  outer arc=3pt,
  boxsep=0pt,
  left=7pt,
  right=7pt,
  top=5pt,
  bottom=5pt,
  lefttitle=7pt,
  righttitle=7pt,
  toptitle=4pt,
  bottomtitle=4pt,
  before skip=6pt,
  after skip=8pt,
  before upper={%
    \setlength{\parindent}{0pt}%
    \setlength{\parskip}{3pt}%
  },
  #1
}

\newenvironment{itemsize*}%
 {\leftmargini=20pt\begin{itemize}%
  \setlength{\itemsep}{3pt}%
  \setlength{\parskip}{0pt}%
  }%
 {\end{itemize}}
\newenvironment{enumerate*}%
 {\begin{enumerate}%
  \setlength{\itemsep}{0pt}%
  \setlength{\parskip}{0pt}}%
 {\end{enumerate}}

\usepackage{blindtext} % to quickly get a full document
\usepackage[normalem]{ulem}

\usepackage{xparse}
\usepackage{colortbl}

\usepackage{svg}

\newcommand{\institutionlogos}{%
  \raisebox{-0.5\height}{\includegraphics[height=0.85cm,viewport=0 1 780 705,clip]{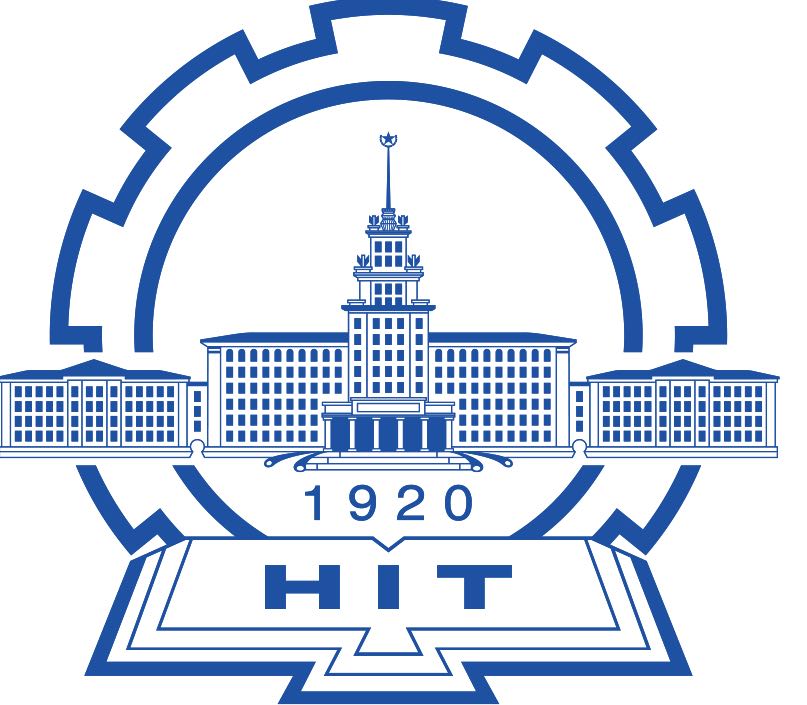}}%
  \kern16pt%
  \raisebox{-0.5\height}{\includegraphics[height=0.85cm,viewport=0 0 32.4 29.04,clip]{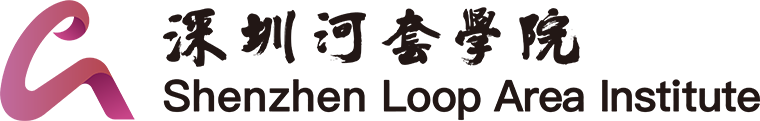}}%
  \kern16pt%
  \raisebox{-0.5\height}{\includegraphics[height=0.85cm,viewport=124.03 25.01 432.10 333.08,clip]{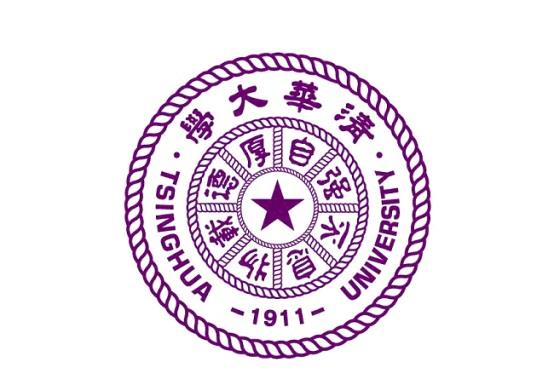}}%
  \kern16pt%
  \raisebox{-0.5\height}{\includegraphics[height=0.85cm,viewport=0 2 578 578,clip]{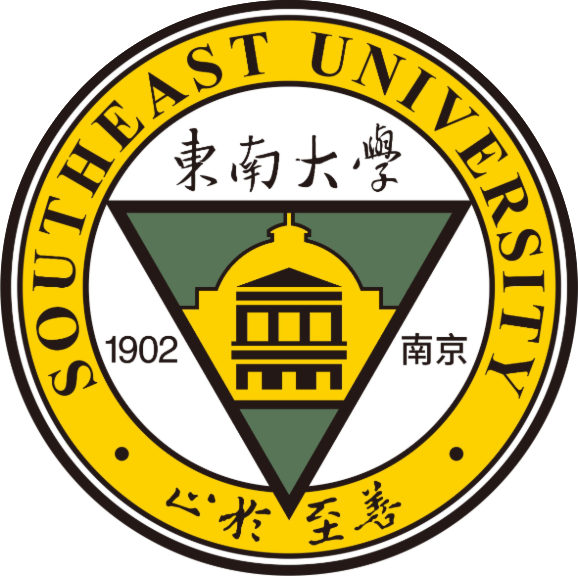}}%
  \kern16pt%
  \raisebox{-0.5\height}{\includegraphics[height=0.85cm,viewport=0 0 48 46,clip]{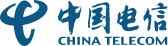}}%
  \kern16pt%
  \raisebox{-0.5\height}{\includegraphics[height=0.85cm]{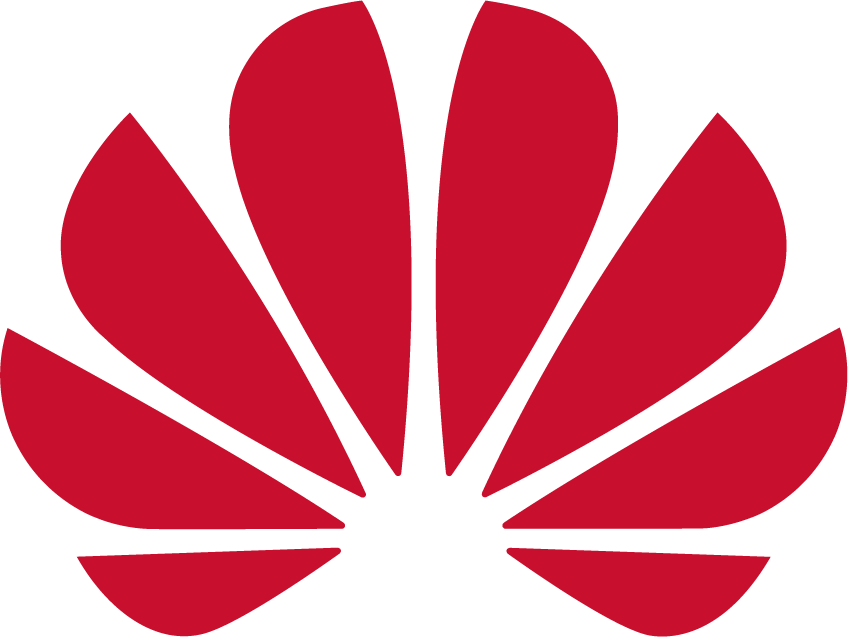}}%
}

\title{Memory-Centric Continual Learning for AI Agents}
\hypersetup{pdftitle={Memory-Centric Continual Learning for AI Agents}}
\author{\institutionlogos}

\begin{document}

\maketitle

\input{sections/abstract}
\input{sections/frontpage-scaling}

\clearpage
{
  \hypersetup{linkcolor=RoyalBlue, linktoc=page}
  \setcounter{tocdepth}{2}
  \tableofcontents
}

\newpage

\input{sections/introduction}
\input{sections/taxonomy}
\input{sections/lifecycle}
\input{sections/evaluation}

\input{sections/applications}
\input{sections/challenges}
\input{sections/conclusion}

% \newpage

\bibliographystyle{lychee}
\bibliography{custom}
\input{sections/appendix}
\end{document}

%% file: sections/abstract.tex
\begin{abstract}
Autonomous agents powered by foundation models can reason, plan, and use tools, but their fixed weights and context limits hinder sustained learning from experience over long-term deployment. While external memory systems attempt to bridge this gap, they mostly act as passive retrieval caches, evaluated primarily by whether they can recall stored facts or past interactions. Whether memory can drive genuine continual learning remains an open question. In this paper, we introduce \textit{Memory-Centric Continual Learning} (MCCL), arguing that memory should be judged by whether it improves future actions, not merely what it retrieves. MCCL has three main parts. First, a three-axis taxonomy organizes memory by where state is stored (\textit{substrates}), what form it takes (\textit{assets}), and how it is managed (\textit{policies}). This defines a two-loop lifecycle that updates memory state in an inner loop while refining management policies in an outer loop based on task feedback. Second, an evaluation protocol measures retention under interference, adaptation to change, and task transfer to isolate genuine behavioral gains from simple context expansion. Third, a memory scaling framework models how agent performance scales with experience, storage capacity, and compute budgets across non-parametric and parametric memory. Together, these components reframe agent memory from passive lookup to continuous learning, offering a principled route toward self-improving autonomous agents.
\end{abstract}

%% file: sections/frontpage-scaling.tex
% Vector excerpts from the Section 4.3 theoretical figures.
% Viewports remove only adjacent panels and the original panel titles.
\begin{figure}[H]
  \centering
  \vspace{7pt}
  \captionsetup{font=small,skip=7pt}
  \begin{minipage}[t]{0.425\linewidth}
    \centering
    {\small\bfseries Non-parametric memory\par}
    \vspace{6pt}
    \includegraphics[width=\linewidth,
      viewport=0 168 253 312,clip]{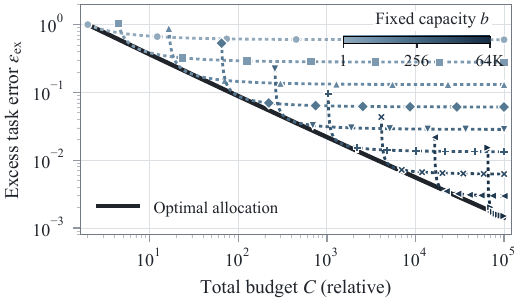}
  \end{minipage}\hfill
  \begin{minipage}[t]{0.555\linewidth}
    \centering
    {\small\bfseries Parametric memory\par}
    \vspace{6pt}
    \includegraphics[width=\linewidth,
      viewport=0 0 343 158,clip]{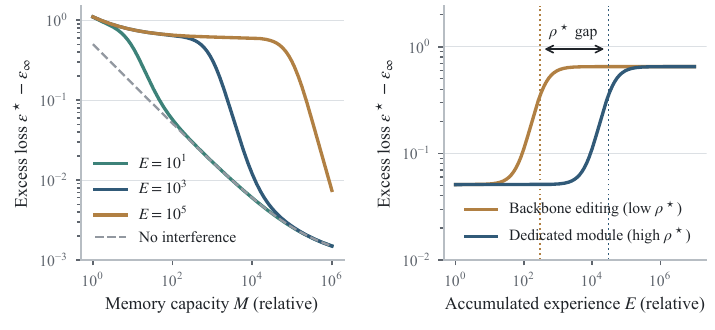}
  \end{minipage}
  \caption{\textbf{Memory scaling across two substrates.} Theoretical illustrations:
  \textit{left}, joint allocation of storage and reading resources sustains improvement
  beyond fixed-capacity limits; \textit{center}, greater experience requires more
  parametric memory capacity; \textit{right}, load-induced interference emerges at
  substrate-dependent thresholds. Curves are synthetic illustrations;
  see Section~\ref{sec:memory-scaling} and Figures~\ref{fig:ms-law} and~\ref{fig:pm-law}.}
  \label{fig:frontpage-scaling}
\end{figure}

%% file: sections/introduction.tex
\section{The Landscape: Why Agents Need Continual Learning via Memory}
\label{sec:introduction}

Autonomous agents powered by foundation models have demonstrated strong
capabilities in multi-step reasoning, planning, and tool use across digital and physical
domains~\cite{agashe2024agents,wang2023voyager,yao2023react}.
However, current agents cannot reliably improve from experience over long deployments.
Although modern agent harnesses can systematically track execution logs, timestamps,
and environment states, they often treat accumulated history as raw text to be retrieved
and inserted back into the prompt.
Simply retrieving raw text does not guarantee behavioral improvement. As interaction
trajectories grow, unmanaged context dilutes relevant signals, increases inference latency,
and degrades model attention~\cite{wang2026evomembench}.

Conventional adaptation methods face clear trade-offs. Expanding context windows is bounded
by session limits and suffers from attention degradation over long contexts. Meanwhile, continual
fine-tuning requires substantial computation, risks catastrophic forgetting of general abilities,
and lacks reliable mechanisms to roll back errors~\cite{gu2024editingharms,luo2023catastrophic}.
Without mechanisms to distill, update, and transfer accumulated experience, deployed agents cannot
retain what they learn. Consequently, they repeatedly commit the same errors, struggle with
conflicting or outdated information, and fail to improve across sequential tasks~\cite{cattan2025conflicts,hu2026memorycl,yu2026forget}.

Agent memory has emerged as a primary approach to support persistence across sessions and
personalization~\cite{chhikara2025mem0,packer2023memgpt,park2023generative}.
However, despite rapid progress, most existing methods focus on specific engineering implementations
rather than systematic continual learning. Many approaches adopt an information retrieval view,
treating memory simply as an external database or a Retrieval-Augmented Generation (RAG) cache
designed to feed past tokens back into the prompt~\cite{lewis2020rag,luo2026storage}.
Furthermore, existing classifications often blur distinct system choices, including where memory is
physically stored, what form knowledge takes, and how memory operations are managed.
Evaluation also remains limited. Many benchmarks primarily test whether an agent can recall or answer
questions over a static past history, rather than whether experience accumulated during deployment
improves subsequent actions under interference and drift~\cite{du2024perltqa,maharana2024locomo,wu2025longmemeval}.
Moreover, as memory stores grow, unmanaged external memory suffers from retrieval noise, memory
pollution, and negative transfer when task distributions shift~\cite{chen2026experienceinternalization,zhang2026faultymemories}.

To bridge this gap, we propose the framework of \textbf{Memory-Centric Continual Learning} (MCCL).
We argue that addressing the stability--plasticity dilemma does not strictly require continuous
gradient updates to foundation model weights.
Instead, memory itself can serve as an evolving learning infrastructure through which an agent
incorporates feedback, refines strategies, and resolves knowledge conflicts during deployment,
while keeping the base model intact.
In this framework, continual learning operates as a coupled dynamical system. An inner
\emph{memory-state loop} updates stored knowledge from sequential deployment experience, while an
outer \emph{policy-adaptation loop} refines memory management policies based on task performance.
By connecting task outcomes back to memory management, MCCL provides a principled route toward
\textbf{Recursive Self-Improvement (RSI)}, where agents not only accumulate experience but also
improve how that experience is organized and used. We summarize our core contributions as follows.

\begin{enumerate}
    \item \textbf{A unified taxonomy of memory architecture.}
    We clarify existing classifications through a three-axis taxonomy:
    \emph{where} experience persists (\textbf{substrates}: contextual, activation, external addressable, and adaptive parametric),
    \emph{what} form it takes (\textbf{assets}: episodic, semantic, and procedural), and
    \emph{how} operations are managed (\textbf{policies}: prescribed, deliberative, and learned).
    This separates physical storage from cognitive representation and operational control
    (Section~\ref{sec:taxonomy}).

    \item \textbf{The memory lifecycle as an engine for continual learning.}
    We formalize memory dynamics across five core stages: selective formation, structural consolidation,
    reactivation and utilization, knowledge revision and selective forgetting, and feedback-driven policy evolution.
    This formulation explains how raw interactions turn into generalizable knowledge while mitigating memory pollution and drift
    (Section~\ref{sec:memory_lifecycle}).

    \item \textbf{A behavioral evaluation framework.}
    We distinguish retrieval-level checks from genuine continual learning outcomes.
    We introduce a four-gate retention cascade (preservation, access, utilization, and effect) to diagnose failure points,
    use matched stateful-versus-reset comparisons ($\Delta P_t$) to isolate the effect of accumulated memory,
    and establish evaluation protocols to measure retention under interference, adaptation under drift, and cross-task transfer
    (Section~\ref{sec:evaluation}).

    \item \textbf{A memory scaling law.}
    We model how agent performance scales with experience, storage capacity, and compute budgets across both non-parametric and parametric memory.
    Both share a resource-limited scaling frontier, but differ in how new writes interact with existing content.
    This shifts the focus from whether larger memory helps to how experience, capacity, and compute should scale together
    (Section~\ref{sec:memory-scaling}).
\end{enumerate}

In summary, this paper reframes agent memory from a passive storage component into an active learning
infrastructure, focusing on whether an agent can continually learn over time rather than simply retrieve
past records. The remainder of this paper is organized as follows.
Section~\ref{sec:taxonomy} presents the taxonomy across substrates, assets, and policies.
Section~\ref{sec:memory_lifecycle} details the memory lifecycle as a continual learning engine.
Section~\ref{sec:evaluation} formalizes the evaluation framework and memory scaling laws.
Section~\ref{sec:applications} analyzes applications across digital and physical domains.
Section~\ref{sec:challenges} discusses open challenges in credit assignment, parametric synergy, and trustworthy memory governance.
Section~\ref{sec:conclusion} concludes the paper.

%% file: sections/taxonomy.tex
\begin{figure}[t]
    \centering
    \includegraphics[width=\linewidth]{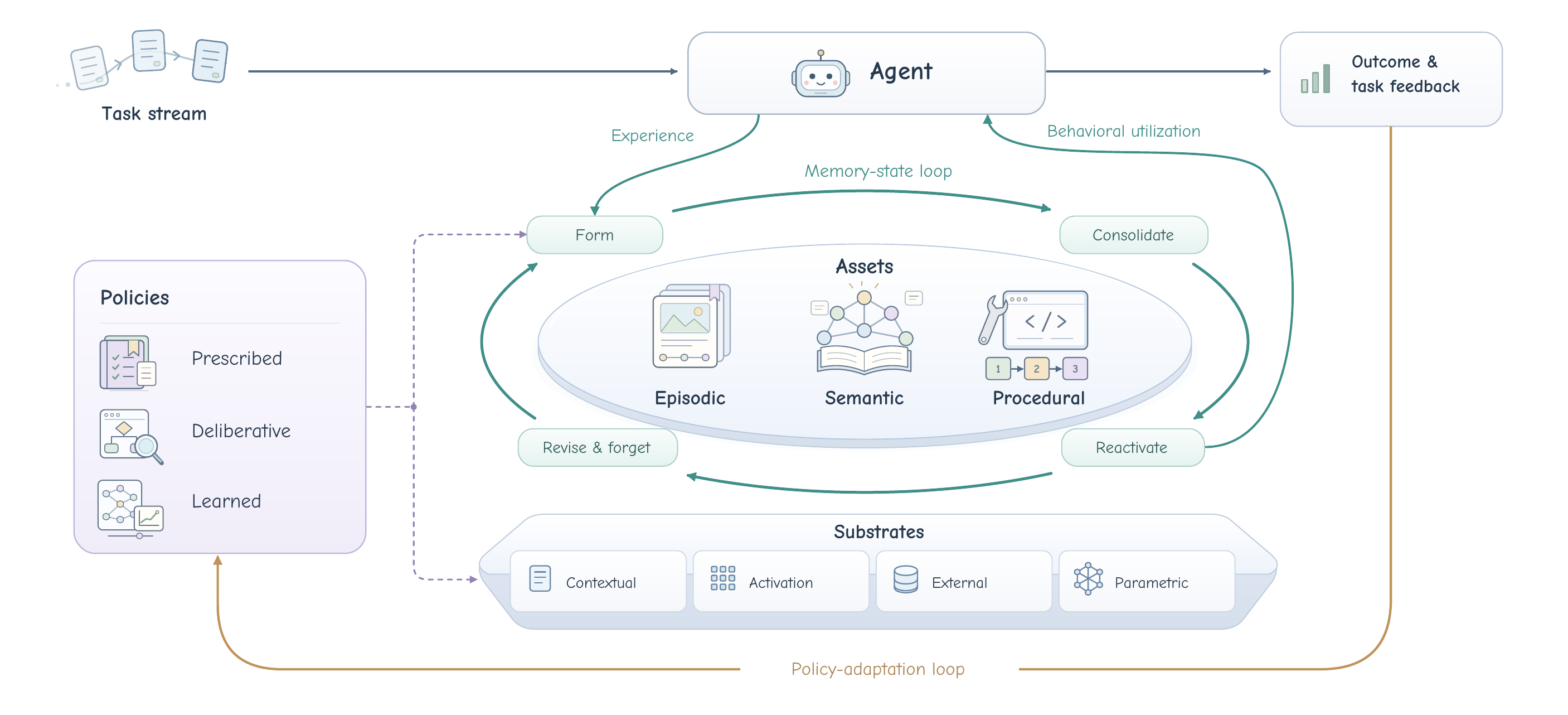}
    \caption{Memory-centric continual learning as a coupled dynamical system.
    A task stream is executed by the agent and produces outcome feedback.
    Experience is organized as episodic, semantic, and procedural assets that
    persist across contextual, activation, external, and parametric
    substrates. An inner memory-state loop forms, consolidates, reactivates,
    and revises these assets, while prescribed, deliberative, or learned
    policies govern the corresponding operations. An outer
    policy-adaptation loop uses task feedback to update memory management.}
    \label{fig:agent-memory-taxonomy}
\end{figure}

\section{A Unified Framework: Architecting Memory for Continual Learning}
\label{sec:taxonomy}

Classical continual learning theory has primarily focused on the
stability--plasticity dilemma within model weights, seeking
to mitigate catastrophic forgetting during sequential gradient
updates~\cite{mccloskey1989catastrophic,wang2024surveycl}.
In contrast, foundation-model-based agents decouple computation from
persistent storage, distributing experiential adaptation across physical
media, semantic abstractions, and control mechanisms.
However, existing taxonomies often mix these levels, confusing where an experience
is stored with what form it takes and how it is managed~\cite{huang2026agentmemory,zhoubian2026memory}.
To establish a principled foundation for memory-centric continual learning,
we organize the design space around three core axes, covering
\emph{where} experience persists, \emph{what} reusable form it takes,
and \emph{how} memory operations are managed
(Figure~\ref{fig:agent-memory-taxonomy}).

\begin{itemize}
    \setlength{\itemsep}{2pt}
    \setlength{\parsep}{0pt}
    \setlength{\topsep}{2pt}
    \interlinepenalty=10000
    \clubpenalty=10000
    \widowpenalty=10000
    \item \textbf{Memory Substrates (WHERE): Physical Storage Media.}
    Substrates define the physical medium where experience persists. They govern trade-offs
    between write latency, volatility, capacity, and durability, determining whether
    experience is held in active context, model activations, external stores, or model parameters.

    \item \textbf{Memory Assets (WHAT): Cognitive Representations.}
    Assets describe the form knowledge takes. Independent of physical substrates,
    assets balance specific detail against general abstraction, grouping experience into
    episodic traces, semantic facts, or procedural skills.

    \item \textbf{Memory Policies (HOW): Mechanisms for Memory Management.}
    Policies define the rules for writing, consolidating, retrieving, updating, and
    forgetting memories. They span fixed heuristics (prescribed), runtime language model
    reasoning (deliberative), and feedback-driven control (learned), determining whether
    memory management remains static or adapts over time.
\end{itemize}

These three axes are complementary dimensions rather than isolated categories.
A concrete agent memory system combines specific choices across all three axes,
such as compiling episodic traces into an external addressable store governed by
deliberative policies powered by Large Language Models (LLMs).
Decoupling these dimensions clarifies that possessing an external substrate
does not by itself yield transferable capability, nor does storing an
episodic asset indicate that an agent has an effective policy to retrieve
and use it.
By formalizing this three-axis architecture, we establish a principled framework
to diagnose where existing architectures succeed or fail under non-stationary streams,
and to compose hybrid systems that support RSI.

\subsection{Memory Substrates: Physical Foundations for Stability and Plasticity}
\label{subsec:2.1}
Memory substrates define where experience persists and how it can be accessed
and modified. Existing taxonomies commonly separate internal from external memory,
or computation-coupled from addressable memory~\cite{huang2026agentmemory,zhoubian2026memory}.
We distinguish substrates by where state physically persists, rather than how it appears
during inference. For example, text retrieved from a database remains external memory even when
placed in a prompt; conversely, offloading the KV cache to disk changes its storage location
but keeps it an activation-level representation. Following this principle, we identify
four basic substrates: contextual memory, activation memory, external addressable memory,
and adaptive parametric memory.

This distinction is independent of memory assets and policies. A substrate
specifies how long experience can persist, how it can be addressed, and at
what granularity it can be updated. It does not, by itself, tell us whether
those updates yield transferable improvement; that depends on the assets
built from experience and on the policies that govern their formation and
use.

\paragraph{Contextual memory.}
The most direct memory substrate is the sequence of tokens currently visible
to the model. Thoughts, actions, and observations can be recorded in a shared
prompt so that the trajectory influences subsequent decisions through
self-attention~\cite{yao2023react}. Ever-growing traces are replaced with
subgoal-level summaries or compressed reasoning
states~\cite{hu2024hiagent,wu2025resum}, and deletion, compression, and
rewriting are exposed as agent actions~\cite{zhang2025memoryaction}. These
methods require little architectural modification and offer nearly immediate
plasticity, with relatively high fidelity over short horizons. Their
stability, however, is bounded by the session limit. As trajectories grow,
older observations consume context tokens and degrade model attention.
Even within the supported context window, performance can drop sharply with
sequence length, evidence position, and task complexity~\cite{liu2023lostmiddle,hsieh2024ruler}.
Summarization and compaction extend the effective horizon but introduce an irreversible,
often lossy selection step~\cite{cim2026parallelcompaction}.

\paragraph{Activation memory.}
A second family compresses history into machine-native states, including
hidden activations, KV caches, recurrent states, and dedicated memory slots.
Hidden states are reused across segments, and KV entries are selectively
retained under a fixed budget~\cite{dai2019transformerxl,zhang2023h2o,liu2025chunkkv}. Previously computed
internal representations are placed in a retrievable memory
bank~\cite{wang2023longmem,wu2022memorizing}, and latent pools or designated
slots are introduced that can be repeatedly read and
updated~\cite{wang2024memoryllm,kang2025lm2}. Compared with textual context,
these states participate in the forward pass at higher bandwidth and need not
be translated back into natural language. Gating, state transitions, and
isolated slots can also confine rapid adaptation to a restricted state
channel, leaving the pretrained backbone unchanged. This efficiency shifts
the capacity cost to runtime memory and limits controllability: activation
memories are tied to a particular model, layer structure, and representation
space, and their contents are difficult to inspect, edit at the level of
individual facts, or transfer across models~\cite{kwon2023pagedattention}.
Fixed-dimensional latent states also face capacity limits as experience
accumulates. Offloading activation states or the KV cache to disk does not turn them
into external text memory, because they can still only be processed as internal
representations by the original model.

\paragraph{External addressable memory.}
External memory separates storage from model computation. Experience persists
as independently maintained entries that can be explicitly retrieved and
modified. Implementations range from natural-language logs and summaries to
vector indices, relational tables, knowledge graphs, hierarchical trees, code
repositories, and collections of specialized stores. Observation streams are
combined with higher-order reflections, language-level diagnoses are retained
from failed trials, and information is moved between a bounded active context
and larger archival
stores~\cite{park2023generative,shinn2023reflexion,packer2023memgpt}. Memory
extraction, dynamic organization, and structured retrieval are further
developed in recent
systems~\cite{chhikara2025mem0,xu2025amem,gutierrez2025hipporag2,rezazadeh2024memtree}.
This substrate offers a direct route to non-parametric
continual learning: new experience can survive across sessions without
altering the foundation model, individual records can be revised or
versioned, and capacity can grow independently of model size.

Structural separation changes rather than resolves the
stability--plasticity dilemma. External stores avoid catastrophic forgetting
caused by parameter updates, but an old memory can become inaccessible when
its retrieval rank declines, producing a form of \emph{functional
forgetting}. Summarization, deduplication, and conflict resolution may also
overwrite earlier records or change whether they still count as valid
evidence. As the store grows, redundant, stale, and conflicting entries
raise retrieval noise. Indexing, provenance tracking, and deletion become
more expensive as well. Irrelevant or
conflicting retrieval can be actively
harmful~\cite{wang2024astuterag,cattan2025conflicts}, and current memory
agents do not reliably combine retrieval, online updating, long-range use,
and selective forgetting~\cite{hu2026memoryagentbench,uddin2026memora}.

\paragraph{Adaptive parametric memory.}
Parametric substrates encode deployment experience in model weights,
detachable adapters, or designated fast parameters. Static pretrained weights
are not sufficient for this category; the defining property is that
parameters can still be updated from experience after deployment. Consecutive
edits are isolated in a side parameter space, and behavior obtained with
visible context is distilled into a model that no longer receives that
context~\cite{wang2024wise,wang2024selfparam}. Test-time neural memory and
rapidly updated LoRA parameters are used to preserve behavioral effects after
the corresponding evidence has left the active
prompt~\cite{behrouz2025titans,ren2026tmem}. Once internalized, it can be used
during a standard forward pass without external retrieval
and may generalize beyond the specific instance that produced the update. It
is therefore attractive for stable, frequently reused, and well-validated
knowledge or skills.

This efficiency at read time comes at the cost of more expensive and less
controllable writes. Parameter updates require optimization, generated
supervision, or specialized test-time learning rules. Once an experience is
distributed across weights, its scope is difficult to localize, and
record-level expiration, provenance tracking, and deletion become
substantially harder than in an external store. Repeated adaptation can cause
catastrophic forgetting of earlier knowledge or consolidate transient
feedback and self-generated errors into persistent behavior. Continual
instruction tuning forgets across domain knowledge, reasoning, and reading
comprehension, and successful factual edits can still degrade unrelated
general abilities~\cite{luo2023catastrophic,gu2024editingharms}. Side
memories, low-rank adapters, and slow--fast parameter decompositions reduce
forgetting by restricting the update scope, but introduce routing failures,
module-merging conflicts, and long-term drift of their own.

\paragraph{Hybrid substrates.}
Practical memory systems increasingly combine substrates with complementary roles,
where tokens and activations support immediate adaptation, external stores provide durable
and auditable retention, and parameter updates selectively internalize stable skills~\cite{packer2023memgpt,wang2023longmem,behrouz2025titans,ren2026tmem}.
Existing hybrid systems still leave several design choices open, including where new
experience first enters, when it moves to a more durable store, how queries are routed,
which state is authoritative during conflicts, and where erroneous updates can be rolled back.

\begin{figure}[!t]
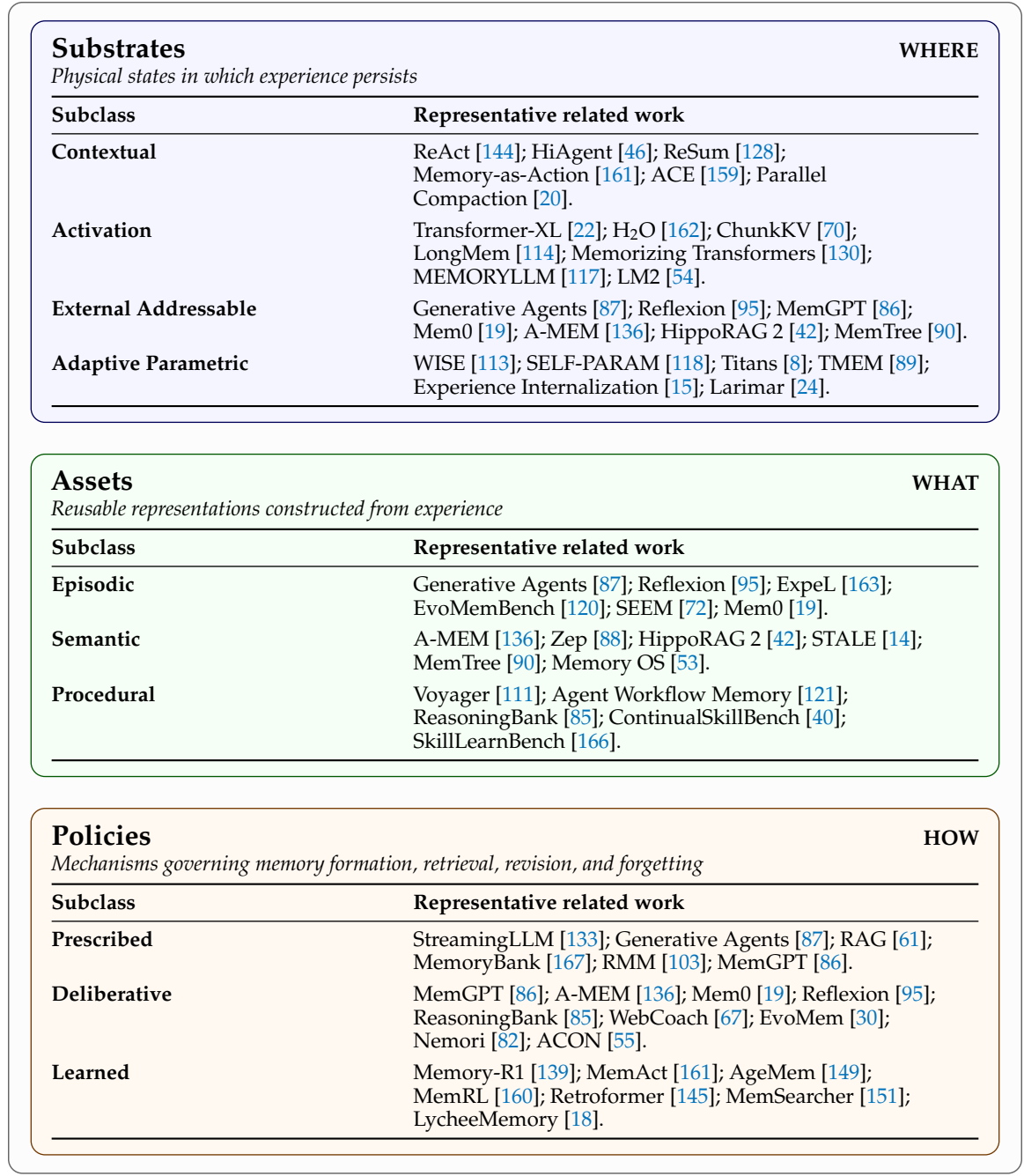

\centering
\small
\setlength{\abovecaptionskip}{6pt}
\setlength{\belowcaptionskip}{0pt}

\begin{tcolorbox}[
    width=\linewidth,
    colback=black!1,
    colframe=black!60,
    boxrule=0.55pt,
    arc=2mm,
    left=2.2mm,
    right=2.2mm,
    top=1.6mm,
    bottom=1.6mm
]

% =========================================================
% SUBSTRATES
% =========================================================
\begin{tcolorbox}[
    width=\linewidth,
    colback=blue!4,
    colframe=blue!35!black,
    boxrule=0.5pt,
    arc=2mm,
    left=2.0mm,
    right=2.0mm,
    top=1.3mm,
    bottom=1.2mm
]

\textbf{\large Substrates}\hfill\textbf{WHERE}\par\vspace{1pt}
{\footnotesize\itshape
Physical states in which experience persists
}

\vspace{0.7mm}

\renewcommand{\arraystretch}{1.05}
\begin{tabularx}{\linewidth}{@{}>{\raggedright\arraybackslash}p{0.36\linewidth}>{\raggedright\arraybackslash}X@{}}
\toprule
\textbf{Subclass} & \textbf{Representative related work} \\
\midrule

\textbf{Contextual}
&
{\footnotesize
ReAct~\cite{yao2023react};
HiAgent~\cite{hu2024hiagent};
ReSum~\cite{wu2025resum};
Memory-as-Action~\cite{zhang2025memoryaction};
ACE~\cite{zhang2025ace};
Parallel Compaction~\cite{cim2026parallelcompaction}.
}
\\ \addlinespace[3pt]

\textbf{Activation}
&
{\footnotesize
Transformer-XL~\cite{dai2019transformerxl};
H$_2$O~\cite{zhang2023h2o};
ChunkKV~\cite{liu2025chunkkv};
LongMem~\cite{wang2023longmem};
Memorizing Transformers~\cite{wu2022memorizing};
MEMORYLLM~\cite{wang2024memoryllm};
LM2~\cite{kang2025lm2}.
}
\\ \addlinespace[3pt]

\textbf{External Addressable}
&
{\footnotesize
Generative Agents~\cite{park2023generative};
Reflexion~\cite{shinn2023reflexion};
MemGPT~\cite{packer2023memgpt};
Mem0~\cite{chhikara2025mem0};
A-MEM~\cite{xu2025amem};
HippoRAG~2~\cite{gutierrez2025hipporag2};
MemTree~\cite{rezazadeh2024memtree}.
}
\\ \addlinespace[3pt]

\textbf{Adaptive Parametric}
&
{\footnotesize
WISE~\cite{wang2024wise};
SELF-PARAM~\cite{wang2024selfparam};
Titans~\cite{behrouz2025titans};
TMEM~\cite{ren2026tmem};
Experience Internalization~\cite{chen2026experienceinternalization};
Larimar~\cite{das2024larimar}.
}
\\
\bottomrule
\end{tabularx}

\end{tcolorbox}

\vspace{1.1mm}

% =========================================================
% ASSETS
% =========================================================
\begin{tcolorbox}[
    width=\linewidth,
    colback=green!4,
    colframe=green!35!black,
    boxrule=0.5pt,
    arc=2mm,
    left=2.0mm,
    right=2.0mm,
    top=1.3mm,
    bottom=1.2mm
]

\textbf{\large Assets}\hfill\textbf{WHAT}\par\vspace{1pt}
{\footnotesize\itshape
Reusable representations constructed from experience
}

\vspace{0.7mm}

\renewcommand{\arraystretch}{1.05}
\begin{tabularx}{\linewidth}{@{}>{\raggedright\arraybackslash}p{0.36\linewidth}>{\raggedright\arraybackslash}X@{}}
\toprule
\textbf{Subclass} & \textbf{Representative related work} \\
\midrule

\textbf{Episodic}
&
{\footnotesize
Generative Agents~\cite{park2023generative};
Reflexion~\cite{shinn2023reflexion};
ExpeL~\cite{zhao2024expel};
EvoMemBench~\cite{wang2026evomembench};
SEEM~\cite{lu2026seem};
Mem0~\cite{chhikara2025mem0}.
}
\\ \addlinespace[3pt]

\textbf{Semantic}
&
{\footnotesize
A-MEM~\cite{xu2025amem};
Zep~\cite{rasmussen2025zep};
HippoRAG~2~\cite{gutierrez2025hipporag2};
STALE~\cite{chao2026stale};
MemTree~\cite{rezazadeh2024memtree};
Memory OS~\cite{kang2025memoryos}.
}
\\ \addlinespace[3pt]

\textbf{Procedural}
&
{\footnotesize
Voyager~\cite{wang2023voyager};
Agent Workflow Memory~\cite{wang2024awm};
ReasoningBank~\cite{ouyang2025reasoningbank};
ContinualSkillBench~\cite{guan2026continualskillbench};
SkillLearnBench~\cite{zhong2026skilllearnbench}.
}
\\
\bottomrule
\end{tabularx}

\end{tcolorbox}

\vspace{1.1mm}

% =========================================================
% POLICIES
% =========================================================
\begin{tcolorbox}[
    width=\linewidth,
    colback=orange!5,
    colframe=orange!45!black,
    boxrule=0.5pt,
    arc=2mm,
    left=2.0mm,
    right=2.0mm,
    top=1.3mm,
    bottom=1.2mm
]

\textbf{\large Policies}\hfill\textbf{HOW}\par\vspace{1pt}
{\footnotesize\itshape
Mechanisms governing memory formation, retrieval, revision, and forgetting
}

\vspace{0.7mm}

\renewcommand{\arraystretch}{1.05}
\begin{tabularx}{\linewidth}{@{}>{\raggedright\arraybackslash}p{0.36\linewidth}>{\raggedright\arraybackslash}X@{}}
\toprule
\textbf{Subclass} & \textbf{Representative related work} \\
\midrule

\textbf{Prescribed}
&
{\footnotesize
StreamingLLM~\cite{xiao2024streamingllm};
Generative Agents~\cite{park2023generative};
RAG~\cite{lewis2020rag};
MemoryBank~\cite{zhong2024memorybank};
RMM~\cite{tan2025rmm};
MemGPT~\cite{packer2023memgpt}.
}
\\ \addlinespace[3pt]

\textbf{Deliberative}
&
{\footnotesize
MemGPT~\cite{packer2023memgpt};
A-MEM~\cite{xu2025amem};
Mem0~\cite{chhikara2025mem0};
Reflexion~\cite{shinn2023reflexion};
ReasoningBank~\cite{ouyang2025reasoningbank};
WebCoach~\cite{liu2025webcoach};
EvoMem~\cite{fan2025evomem};
Nemori~\cite{nan2025nemori};
ACON~\cite{kang2025acon}.
}
\\ \addlinespace[3pt]

\textbf{Learned}
&
{\footnotesize
Memory-R1~\cite{yan2025memoryr1};
MemAct~\cite{zhang2025memoryaction};
AgeMem~\cite{agemem};
MemRL~\cite{zhang2026memrl};
Retroformer~\cite{yao2024retroformer};
MemSearcher~\cite{yuan2025memsearcher};
LycheeMemory~\cite{chen2026lycheememory}.
}
\\
\bottomrule
\end{tabularx}

\end{tcolorbox}

\end{tcolorbox}

\caption{
\textbf{A unified taxonomy of agent memory for continual learning.}
Memory substrates specify where experience persists, memory assets describe
what reusable representations are constructed from experience, and memory
policies determine how memory is written, retrieved, consolidated, revised,
and forgotten. Representative related works are organized by subclass under
each of the three dimensions.
}

\label{fig:unified-taxonomy}

\end{figure}

\subsection{Memory Assets: Representations for Cross-Task Transfer}

Memory substrates determine the physical state in which experience persists,
whereas memory assets describe the reusable units constructed from that
experience. The two dimensions do not admit a fixed correspondence. The same
task experience may remain temporarily as context tokens, be written to an
external log, organized as a graph, or eventually internalized into model
parameters. Conversely, a single external store may contain event records,
factual relations, and executable skills. We therefore classify an asset by
how it is used when retrieved, rather than by how it is stored, distinguishing
episodic, semantic, and procedural assets. This taxonomy concerns experience
that can persist across interactions or tasks. Contextual tokens and
activation states may participate in memory formation, but they count as
assets only once they persist in a reusable form across tasks; otherwise they
remain working state on a substrate.

As with substrates, possessing an asset is not itself evidence of cross-task
learning. Episodes, facts, and skills specify the \emph{form} in which
experience can transfer; whether transfer occurs is a question for memory
policies.

\paragraph{Episodic assets.}
Episodic assets preserve \emph{what happened}. Their basic unit is a
particular experience situated in its temporal, environmental, and task
context. An episode may include observations, goals, actions, tool calls,
intermediate reasoning, feedback, and outcomes, or a compressed segment
derived from these elements. Compared with more abstract assets, episodes
retain the conditions under which an experience was obtained. They therefore
support case-based reuse, resumption of interrupted tasks, and retrospective
inspection of the evidence behind a later belief or strategy. Observations
and higher-order reflections are stored in a memory stream, trial outcomes
and linguistic diagnoses are retained after failed attempts, and both prior
trajectories and insights extracted from collections of experience are
retrieved for later reuse~\cite{park2023generative,shinn2023reflexion,zhao2024expel}.
Structured event records and production memory stores further retain
temporally grounded episodes across sessions~\cite{lu2026seem,chhikara2025mem0}.

Episodes are particularly valuable for rare events and one-shot observations
for which reliable abstraction is not yet possible. They permit highly
plastic writes because the agent need not decide immediately whether an
observation expresses a general rule, while their high fidelity supplies
evidence for later correction. The same fidelity, however, produces
substantial storage and retrieval costs. As the collection grows,
trajectories that are superficially similar but decisionally different
compete for retrieval, while complete traces carry observations and actions
irrelevant to the current problem. Stored experience is most useful when its
structure matches the target task; in other settings, long-context baselines
remain competitive~\cite{wang2026evomembench}.

\paragraph{Semantic assets.}
Semantic assets preserve \emph{what is known}. Their reusable units are facts,
concepts, relations, constraints, preferences, and environment models
abstracted from particular observations. Unlike episodic assets, they suppress
much of the temporal and surface detail of the originating event so that
knowledge can be reused across sessions and tasks. A recurring user
preference, a tool authentication requirement, or a relation between two
entities can thus be invoked without replaying the full interaction from which
it was inferred.

Stripping temporal and situational detail yields compression and broader
reuse, and structured semantic assets additionally support entity-level
revision, relational inference, and multi-hop retrieval. Dynamically indexed
notes are
linked so that new entries can modify the contextual representation of
existing ones; current and historical relations are maintained in temporal
knowledge graphs; and structured retrieval is framed as non-parametric
continual
learning~\cite{xu2025amem,rasmussen2025zep,gutierrez2025hipporag2}.
Hierarchical schemas and operating-system-style stores likewise organize
reusable facts beyond a single
episode~\cite{rezazadeh2024memtree,kang2025memoryos}. These approaches
indicate that semantic memory is not a static knowledge base but a
knowledge state that must be continually formed, connected, and revised.

The central weakness of semantic assets arises from the same abstraction that
enables transfer. Premature consolidation may convert an accidental
correlation into a general claim, while removing context may discard when,
where, and with what confidence that claim is valid. Appending new information
can leave mutually competing facts in memory, whereas overwriting an earlier
entry can destroy historical provenance. Retrieving updated evidence also
does not guarantee recognition of an implicitly invalidated belief or
corresponding adaptation in downstream behavior~\cite{chao2026stale}. In
practice, existing semantic stores often keep the extracted claim while
omitting provenance, temporal scope, confidence, and links to the episodes
that support it.

\paragraph{Procedural assets.}
Procedural assets preserve \emph{how to act}. Their reusable units include
workflows, strategies, plan templates, tool-use rules, and executable
programs. They may be represented as natural-language instructions, code,
parameterized workflows, or behavioral policies internalized in model
parameters. The distinction from semantic assets is functional rather than
syntactic: semantic assets describe a state of the world, whereas procedural
assets prescribe how that state should be changed under particular
conditions. The same piece of text can therefore be semantic if retrieved as
a fact and procedural if retrieved as an instruction.

Because they directly constrain action generation, procedural assets provide
the most immediate form of behavioral transfer. Environment-verified skills
are stored as composable code, parameterized workflows are induced from
successful trajectories for reuse across tasks and domains, and
generalizable strategies are extracted from successful and failed experience
instead of retaining only raw
trajectories~\cite{wang2023voyager,wang2024awm,ouyang2025reasoningbank}.
Evolving context playbooks and learned memory skills further compile
repeated execution into reusable action
knowledge~\cite{zhang2025ace,zhang2026memskill,ni2026trace2skill}. Procedural
assets can also be composed, allowing previously acquired skills to become
components of more complex behavior and reducing repeated exploration.

This compilation nevertheless locks in assumptions from the original
experience. A workflow may remain valid only for a particular tool
version, interface, or set of environment dynamics. Retrieving a skill for a
superficially similar task with different preconditions can produce negative
transfer, while unverified self-reflection may turn an accidental success into
a persistent rule. Explicit skill maintenance is not uniformly superior to
retaining prior context, and gains are more reliable on tasks with clear
workflows than on open-ended ones~\cite{guan2026continualskillbench,zhong2026skilllearnbench}.
In current systems, preconditions, expected outcomes, environment
dependencies, verification status, and versioning are often left implicit.
Long-lived skill libraries therefore tend to accumulate skills rather than
merge, deprecate, or regression-test them.

\paragraph{Cross-asset consolidation.}
The three asset classes are not mutually exclusive storage compartments. They
describe experience at different levels of abstraction. Episodic assets
retain concrete evidence; regularities recurring across episodes may be
consolidated into semantic assets; and repeatedly validated action
regularities may be compiled into procedural assets. This transformation
improves compression and cross-task reuse while progressively removing details
of the original context. Asset formation therefore exposes a fundamental
trade-off: representations close to individual episodes preserve evidence but
depend on precise matching, whereas abstract knowledge and skills are cheaper
to invoke but allow erroneous generalizations to affect a wider range of
tasks.

Consolidation is rarely a clean, one-way pipeline in existing systems.
Semantic claims and procedural skills often lose their links to supporting
episodes, and new experience cannot always qualify, revise, or revoke what
was previously consolidated.
A single record can also contain all three components, such as a task trajectory,
an environmental fact inferred from it, and a corrective procedure learned after failure.
At retrieval time, such a record is better classified by how it is used than by how it is
physically stored.

\subsection{Memory Policies: Prescribed, Deliberative, and Learned Control}

As agents operate over longer horizons, memory is no longer only a problem of
storing more information or retrieving it more accurately. The central
challenge is how memory should change over time, deciding what should be retained,
consolidated, revised, or forgotten, and how these decisions should adapt to
new experience and feedback. We refer to the mechanisms governing such
decisions as memory policies. They determine when and how memory state
evolves, and they correspond to the \emph{how} column of
Figure~\ref{fig:unified-taxonomy}.

We distinguish three control regimes. \emph{Prescribed} policies apply fixed
rules, heuristics, and scoring functions. \emph{Deliberative} policies use
the LLM to decide, for each case, whether to write, retrieve, or revise,
given the task, memory state, and context. \emph{Learned} policies update
memory control from outcomes and feedback. The regimes are compositional
rather than exclusive: a system may prescribe capacity limits, deliberate
over conflicts, and learn retrieval utilities. They also align with the
prescribed, deliberative, and learned control tiers used later in the lifecycle
comparison (Section~\ref{sec:memory_lifecycle}).

\paragraph{Prescribed policies.}
Prescribed policies treat memory management as resource allocation under
limited context, cache, and store capacity. Control is specified before
deployment through rules such as FIFO, LRU, sliding windows, or fixed
Top-$k$ retrieval, so utility is approximated by position, recency, and
budget rather than estimated future value~\cite{xiao2024streamingllm}.
Salience-based scoring is a richer but still prescribed variant: recency,
importance, and relevance are combined to select episodic
records~\cite{park2023generative,lewis2020rag}, and entries are admitted or
forgotten according to novelty, importance, or decay
thresholds~\cite{zhong2024memorybank,tan2025rmm}. Hierarchical designs go
beyond single-shot Top-$k$ by controlling movement across memory tiers, yet
the tools, thresholds, and control rules remain
predefined~\cite{packer2023memgpt}. Prescribed control is efficient,
predictable, and easier to inspect. Its limitation is rigidity: what
counts as novel, important, or disposable is not subsequently revised
according to task outcomes.

\paragraph{Deliberative policies.}
Deliberative policies place the LLM inside the memory-control loop because
numerical scores cannot fully capture the semantic value of an experience.
The model is asked, for each case, whether to store, retrieve,
resolve a conflict, or rewrite, using the current task, memory state, and
context. Handcrafted scoring is replaced by case-by-case judgments, while
the outer tool protocol remains
fixed~\cite{packer2023memgpt,xu2025amem,chhikara2025mem0}. A stronger
deliberative pattern lets experience revise subsequent control in natural
language: verbal post-mortems are stored to steer later attempts, and
successful and failed traces are abstracted into reusable
strategies~\cite{shinn2023reflexion,ouyang2025reasoningbank}. Write
boundaries are tightened, consistency is verified against constraint memory,
or the memory collection itself is reorganized~\cite{liu2025webcoach,fan2025evomem,nan2025nemori,kang2025acon}. The resulting
control is flexible and interpretable, but inference-intensive and dependent
on prompt following. Rewriting a strategy in language still leaves the
controller untrained, as most such systems lack credit assignment
from later task outcomes back to earlier memory decisions.

\paragraph{Learned policies.}
Learned policies treat memory control as an experience-updated decision
problem. The agent may still use standard operations, such as write,
retrieve, compress, revise, and forget, but a trainable controller selects among
them, and those selections improve from outcomes and feedback. Discrete
memory actions are learned from
reward~\cite{yan2025memoryr1,zhang2025memoryaction,agemem}, and purely
similarity-based retrieval is replaced with utility values that reflect
historical contribution to success~\cite{zhang2026memrl}. Reinforcement
learning provides the credit assignment that static controllers lack, because the true value
of a write or deletion often appears only several steps
later~\cite{yao2024retroformer,yuan2025memsearcher,chen2026lycheememory}.

Two levels of learning should be distinguished. In many existing systems the
controller is optimized offline and then frozen, so memory content continues
to evolve while the operational policy remains static. Recent RSI frameworks
instead attempt rule-level
modification, where the language model is prompted to inspect and revise its own
prompt templates or execution
scaffolding~\cite{zhang2025ace,agrawal2026gepa,yin2024godel,zhang2026dgm}.
Recuris further shows that a Meta-Agent can synthesize validation-gated
updates from execution failures, recursively refining operational rules and
the skill library~\cite{yu2026recuris}. Although reflection-based prompt
adaptation offers substantial semantic flexibility, such methods typically
rely on heuristic textual edits and trial-and-error screening against a
validation set. Under long-horizon
delayed task feedback, they struggle to establish fine-grained temporal
credit assignment and to keep the meta-policy stable. A more
systematic form of continual policy evolution would update the
memory-management policy from ongoing environmental feedback while
suppressing policy drift. The next section turns to how experience-level
rule evolution can be combined with structured closed-loop policy learning
across the lifecycle.

%% file: sections/lifecycle.tex
\section{The Memory Lifecycle as the Engine of Continual Learning}
\label{sec:memory_lifecycle}

The previous section describes agent memory in terms of where experience
persists, what reusable form it takes, and how memory operations are
controlled. These dimensions define the architecture of a memory system, but
they do not explain how that system changes through continued deployment.
Continual learning is often described as adaptation through parameter updates.
For agents, however, experience can also accumulate in external records,
persistent internal states, and other writable memory substrates. We therefore
view memory-centric continual learning (MCCL) as a coupled dynamical process,
where experience updates the persistent memory state while behavioral feedback
can also adapt the policy governing future memory decisions.

At the first level, newly acquired experience changes the memory state through
a transition function controlled by the current memory policy,
\begin{equation}
    \mathcal{M}_{t+1}
    =
    \mathcal{T}
    \left(
        \mathcal{M}_t,
        e_t;
        \pi_t^{\mathrm{mem}}
    \right),
    \label{eq:memory_state_dynamics}
\end{equation}
where $\mathcal{M}_t$ denotes the persistent memory state at time $t$,
$e_t$ denotes experience from the current interaction, and
$\pi_t^{\mathrm{mem}}$ denotes the policy that controls how this experience
is handled. We use $e_t$ broadly to include relevant observations, agent
actions, outcomes, and feedback. The transition function $\mathcal{T}$ is
independent of any particular memory substrate. Depending on the system, it
may admit new experience, reorganize accumulated memory, revise existing
content, or reduce the future influence of selected memories.

A change in memory state is not, by itself, evidence of continual learning.
The updated state must later be accessed and incorporated into reasoning,
decisions, or actions under the memory policy,
\[
    \mathcal{M}_{t+1}
    \xrightarrow{\text{access and utilization }(\pi_{t+k}^{\mathrm{mem}})}
    \Delta \mathcal{B}_{t+k},
    \qquad k > 0,
\]
where $\Delta \mathcal{B}_{t+k}$ denotes a downstream behavioral effect
associated with the memory change. Information may persist without being
retrieved, may be retrieved without being used appropriately, or may influence
behavior in a harmful way. Memory supports continual learning only when
experience-induced changes remain usable, yield appropriate improvement or
adaptation in later behavior, and avoid unnecessary damage to still-useful
capabilities~\cite{shu2026agentcl,yu2026forget}.

At the second level, feedback may change not only memory content but also the
policy that controls future memory decisions. We write this outer update as
\begin{equation}
    \pi_{t+1}^{\mathrm{mem}}
    =
    \mathcal{U}
    \left(
        \pi_t^{\mathrm{mem}},
        \mathcal{F}_{t:t+H}
    \right),
    \label{eq:memory_policy_dynamics}
\end{equation}
where $\mathcal{F}_{t:t+H}$ summarizes downstream feedback observed over a
later horizon $H$, and $\mathcal{U}$ denotes the rule used to update the
memory policy. Such feedback may include task outcomes, rewards, errors, user
corrections, or other signals that reveal the consequences of earlier memory
decisions. $\mathcal{U}$ need not change the policy in every
system. A fixed controller may continue to modify $\mathcal{M}_t$ while
$\pi_t^{\mathrm{mem}}$ remains unchanged. Policy evolution refers to the
stronger case in which accumulated experience also changes how future memory
decisions are made.

Equations~\eqref{eq:memory_state_dynamics} and
\eqref{eq:memory_policy_dynamics}, together with behavioral utilization,
define two coupled processes. The inner memory-state loop governs both how
the persistent state $\mathcal{M}_t$ evolves and how it informs decisions:
how experience enters persistent memory (\emph{formation and writing}), how
accumulated experience is restructured (\emph{organization and consolidation}),
how prior memory returns to the active decision process (\emph{retrieval and
utilization}), and how the stored state is maintained as new evidence arrives
(\emph{revision and forgetting}). Persistent memory is not an additional
operational stage; rather, it is the central evolving state upon which write,
consolidation, readout, and maintenance operations act. The outer control loop
determines how decisions within these four stages are governed and whether
downstream feedback modifies those control rules over time.

\begin{figure}[t]
    \centering
    \includegraphics[width=0.95\linewidth]{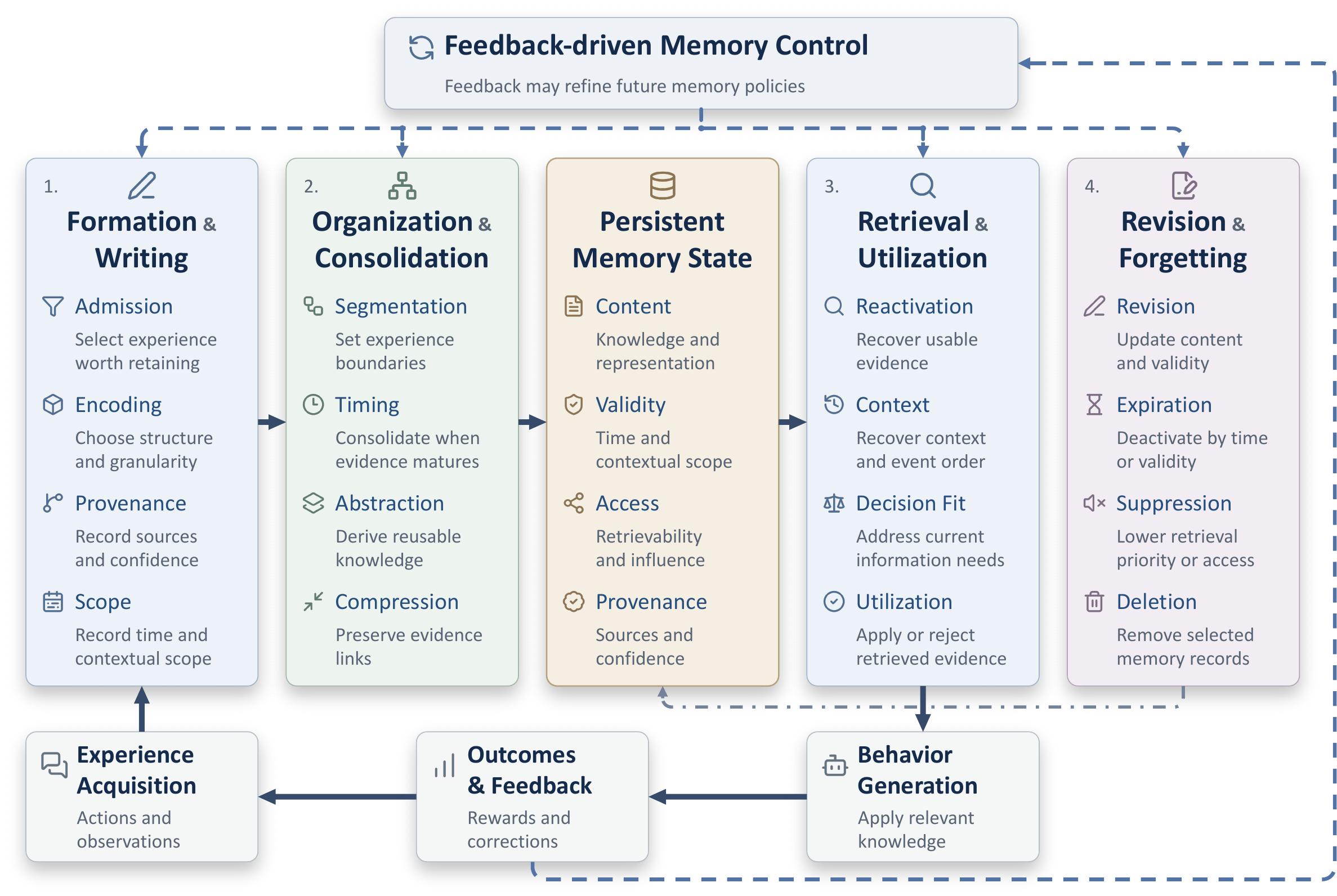}
    \caption{
        The memory lifecycle as two coupled processes in memory-centric
        continual learning. The memory-state loop governs how experience
        enters persistent memory, is organized and reused, and is revised or
        forgotten as conditions change. The control loop uses behavioral
        outcomes to influence future memory decisions.
    }
    \label{fig:memory_lifecycle}
\end{figure}

Together, the two loops describe continual learning as more than memory
accumulation. Experience changes memory, memory influences later behavior,
and behavioral consequences may in turn improve how subsequent experience is
managed. The goal is therefore not to maximize the amount of stored
information or the frequency of updates, but to support adaptation while
preserving valid evidence and useful capabilities.
Figure~\ref{fig:memory_lifecycle} summarizes this view.

\subsection{Formation and Selective Writing}
\label{subsec:formation}
Formation is the first point at which ongoing experience can acquire
persistent influence over future behavior. It determines not only which
observations should enter long-term memory, but also how they are represented
and what source, temporal, and contextual information is preserved for later
use.

\paragraph{Selective Writing and Plasticity Control.}
Memory-centric continual learning begins with an admission decision, because not
every interaction should modify persistent memory. Writing too
freely can accumulate incidental, redundant, or erroneous observations and
increase interference over time. Writing too conservatively can discard
experience that would have supported later adaptation. Selective writing
therefore defines the agent's plasticity boundary by deciding which
deployment-time experiences may exert persistent influence on future
behavior. Selected session-level information can be promoted into persistent
memory, and storage operations can be exposed as actions that are optimized
from downstream outcomes~\cite{cognee,agemem}. These examples
show that admission can range from staged promotion to learned action
selection. In either case, the cost of a poor write can persist, as admitting
unreliable experience propagates errors while rejecting useful experience
removes evidence needed for future learning. The objective is
therefore not to store as much as possible, but to admit useful experience
without giving transient or unreliable observations unnecessary long-term
influence.

\paragraph{Initial Representation.}
Admission also requires a choice about how experience is represented. What is
made explicit at write time determines what can later be retrieved, compared,
or revised. An experience may remain as a relatively direct episodic record
or be encoded into a representation that makes events, entities, relations,
and temporal information explicit~\cite{lu2026seem,rasmussen2025zep}.
Structure and granularity are orthogonal choices: structure determines how
explicitly relational and temporal attributes are indexed, whereas granularity
determines how much situational detail is retained versus compressed. A
representation can therefore be highly structured while remaining fine-grained,
whereas overly coarse or unstructured encodings may discard contextual details
that later become critical. Because subsequent consolidation cannot reliably
recover evidence that was never recorded at admission, the initial
representation directly bounds the evidence available to all downstream memory
operations.

\paragraph{Source, Scope, and Memory Pollution.}
A useful persistent memory should also retain the conditions under which it
was acquired. Relevant information includes provenance, temporal validity,
contextual scope, and, where available, confidence~\cite{memos,rasmussen2025zep}. These attributes do not prove that a memory is true.
Instead, they help later reasoning determine where the memory came from,
whether it still applies, and how strongly it should be trusted. Without such
information, a hallucinated observation may later be reused as fact, an
accidental success may become a general strategy, or a temporary state may be
applied outside the context in which it was observed. This causes memory
pollution, where an error introduced at formation continues to degrade later
behavior and may be reinforced through repeated retrieval, revision, and
consolidation. Formation therefore determines both which experience enters
persistent memory and what evidence is carried forward with it. The next
question is how these formed memories should be organized and transformed as
experience accumulates.

\subsection{Organization, Compression, and Consolidation}
\label{subsec:consolidation}

Organization and consolidation form the integration stage of the memory
lifecycle. Once experience has entered persistent memory, the system must
decide which observations belong together, when accumulated evidence is ready
for further transformation, and whether repeated experience should remain
episodic or become more reusable knowledge. These decisions improve stability
and reuse, but they also introduce risks, as premature or overly aggressive
consolidation can turn incomplete evidence into persistent generalizations.
Here, consolidation refers to the durable organization or transformation of
experience across interactions, rather than the compression of active
short-term context alone.

\paragraph{Choosing the Unit of Experience.}
Consolidation first requires a decision about the unit over which experience
is organized. Continuous interaction arrives as a stream, yet memory systems
must decide which observations should be treated as part of the same event,
episode, or semantic segment. This choice determines which pieces of evidence
can later be compared, linked, or transformed together. Recent systems use
dynamically detected events or temporally grounded episodes instead of relying
only on fixed turns or windows~\cite{esmem,rasmussen2025zep}. Fine-grained units
preserve local detail but can split dependencies across several memories.
Broader units make longer-range relations easier to capture, but they may also
combine observations that should remain separate. Experience boundaries
therefore shape what later consolidation can learn from, determining which
evidence is considered jointly and which patterns can eventually be stabilized
and reused.

\paragraph{When to Consolidate.}
A coherent experience unit does not automatically justify immediate
consolidation. If transformation occurs too early, the system may generalize
from insufficient evidence, repeatedly rewrite unstable knowledge, or amplify
an early mistake. If it occurs too late, related experience remains fragmented
and becomes harder to reuse. Recent approaches make this timing decision
explicit in different ways. LycheeMemory V2 waits for semantic segment
boundaries before encoding accumulated interaction into long-term memory,
whereas RecMem uses recurrence across related observations as a signal that
consolidation is warranted~\cite{li2026lycheememoryv2,recmem}. These methods
treat consolidation timing as an explicit memory-control decision rather than
a routine background process. Their triggers, however, are still prescribed
rather than learned from downstream outcomes. This distinction separates the
question of when evidence is ready for consolidation from the later
policy-level question of whether the rule that makes this decision can itself
improve from feedback.

\paragraph{From Episodes to Reusable Abstractions.}
Consolidation also determines the level at which past experience remains
available for future use. Individual episodes preserve what happened and the
conditions under which it happened, but continual learning often requires
regularities that can be reused beyond a single case. Collections of
trajectories can be distilled into reusable natural-language insights or
executable skills~\cite{zhao2024expel,wang2023voyager}. Evidence from continual
experience internalization, although obtained in a parametric setting,
similarly suggests that principle-level experience may remain more robust
across repeated updates than instance-specific traces~\cite{chen2026experienceinternalization}. Abstraction can therefore improve
transfer, but it should not be treated as a mandatory replacement for
episodic memory. Episodes preserve context and supporting evidence, whereas
more abstract knowledge trades some of that detail for broader applicability.
Consolidation is thus a change in the form and scope of knowledge, not merely
a reduction in storage size. It determines when an experience should remain a
case to be recalled and when repeated evidence is strong enough to support a
more general principle, schema, or procedure.

\paragraph{Lossy Consolidation and Evidence Preservation.}
The same transformation that improves reuse can also remove information that
later becomes important. Repeated LLM-based rewriting has been shown to
initially improve and then reduce memory utility, while retaining raw episodes
can remain competitive with systems that repeatedly force consolidation~\cite{zhang2026faultymemories}. Compression can also remove the contextual
conditions needed to decide whether an abstracted memory still applies.
RaMem describes this failure as context collapse and restores information such
as event time, participants, and session scope before compressed memories are
used as evidence~\cite{ramem}. Consolidation should therefore preserve links
to supporting evidence rather than replace that evidence by default.
Higher-level abstractions should remain traceable to the episodes from which
they were derived so that later retrieval, verification, or revision can
recover from a poor generalization. Consolidation can stabilize and generalize
experience, but it can also become a source of distortion, interference, and
implicit forgetting when abstraction removes information that remains useful.

\subsection{Retrieval and Utilization}
\label{subsec:retrieval_utilization}

Retrieval and utilization (i.e., uptake) connect accumulated experience to the agent's current
reasoning and actions. Their effectiveness depends on recovering sufficient
evidence, matching it to the current decision, and using it appropriately.
These functions concern the use of existing memory rather than changes to
its persistent content.

\paragraph{Reactivating Relevant Evidence.}
Retrieval must recover usable evidence, not simply records with similar
wording. A task may depend on several related events, their temporal order,
or the circumstances in which earlier observations were made. Selecting the
right memory source is therefore only part of the problem. The system must
also recover enough supporting context to interpret what it finds. Temporal
graph memories, reflective memory managers, and contextual-reinstatement
methods offer different ways to select or recover such evidence~\cite{rasmussen2025zep,tan2025rmm,ramem}. These mechanisms connect retrieval to the
representation and evidence-preservation choices discussed above. A compact
summary may be sufficient for one task, whereas another may require access
to its supporting episodes. Recovering that context helps establish the
meaning and scope of the memory without rewriting it. The goal is not to
return as much history as possible, but to recover enough relevant evidence
and to keep its source and context checkable.

\paragraph{Matching Memory to the Decision State.}
A memory can be relevant to a task yet unhelpful at a particular decision
point. During planning, the agent may need a reusable procedure or a record
of earlier constraints. During tool execution, it may instead need exact
parameters or the conditions required for an operation to succeed.
Verification may call for evidence that tests a tentative conclusion, while
failure recovery may benefit from records of unsuccessful attempts.
Retrieval should therefore consider the current subgoal, the evidence
already available, and the uncertainty that prevents the next decision.
This also dictates when retrieval is beneficial, since additional memory may resolve
a specific information gap, repeat existing context, or introduce distracting
details. Relevance to the overall task is therefore
not enough. The retrieved evidence should address what the agent needs to
decide at that moment, rather than simply resemble its current query.

\paragraph{From Access to Effective Use.}
Utilization (often termed \emph{uptake} in retrieval workflows) concerns
whether and how retrieved evidence actually shapes reasoning or action. A
record appearing in the context does not establish that the agent used it, and
its use does not guarantee behavioral improvement. Even after a relevant
record is preserved and retrieved, the agent may overlook it in context,
misinterpret its valid scope, or misapply a past strategy under conditions
where it no longer holds. Studies of continual agents and experience reuse
highlight this gap between retaining experience and benefiting from it
later~\cite{shu2026agentcl,chen2026experienceinternalization}. Consequently,
four sequential conditions must remain conceptually distinct:
\emph{preservation} (retaining the record in storage), \emph{access}
(retrieving it into the decision context), \emph{utilization} (incorporating
it into reasoning), and \emph{behavioral effect} (achieving a net task
improvement, formalized as the retention cascade in
Section~\ref{sec:evaluation}). Appropriate utilization may involve applying a
memory, restricting its use to supported conditions, or rejecting it when
more reliable evidence contradicts it. When memory use reveals an error or outdated assumption that would also harm
future tasks, the system must decide whether and how to modify persistent
memory. That maintenance decision is the subject of revision and forgetting.

\subsection{Revision and Forgetting}
\label{subsec:revision_forgetting}

Revision and forgetting maintain persistent memory as new evidence arrives
and earlier knowledge loses its validity or usefulness. Revision changes
what is recorded or treated as current, while forgetting regulates continued
access and influence. These operations contribute to the state transition in
Equation~\eqref{eq:memory_state_dynamics}. They may be triggered by incoming
experience or by problems discovered during memory use.

\paragraph{Resolving Conflicts and Tracking Current Validity.}
Before modifying memory, the system must determine what the new evidence
changes. A disagreement may indicate an error, a change over time, or a
difference in context. These cases require different responses. Correcting a
false observation is not the same as preserving a historically accurate
observation while marking it as no longer current. Graphiti separates
temporal validity from the retention of historical evidence, while UMA
processes successive observations to maintain an evolving memory state~\cite{rasmussen2025zep,uma}. Source reliability and applicability must also be
considered; newer information should not automatically replace older
evidence. When the available evidence does not resolve a conflict, preserving
the competing records with their sources may be preferable to forcing an
update. The goal is to maintain usable current knowledge without erasing the
conditions under which earlier memories were valid.

\paragraph{Making Knowledge Revision Persistent.}
Once a correction is justified, it should be reflected in persistent memory
so that later tasks do not repeatedly encounter the same inconsistency.
Revision can add missing information, correct an existing record, narrow its
scope, change its validity, or reorganize its links to other memories.
Systems implement these changes through direct updates or through mechanisms
that retain earlier evidence and versions~\cite{uma,vermem,memos}.
These options need not be mutually exclusive. A system can update its current
knowledge while retaining the history needed to inspect or reverse the
change. Revision should also account for derived memories. If an episode
supports a summary or reusable procedure, correcting the episode alone may
leave that derived knowledge inconsistent. Links to supporting evidence help
identify which derived memories may also need review. The objective is not
frequent rewriting, but a justified change whose scope matches the evidence
and whose consequences remain traceable.

\paragraph{Expiration, Suppression, and Deletion.}
Not every maintenance decision concerns factual truth. A memory can remain
accurate yet become redundant, rarely useful, or overly costly to keep active.
Forgetting regulates whether and how such records continue to influence later
behavior. Hard deletion permanently removes a record from its persistent
store; expiration deactivates it according to a predefined retention window
or inactivity schedule; and soft suppression retains the record while
lowering its retrieval priority or accessibility. MemoryBank uses
time-dependent memory strength decay, whereas VerMem supports soft deletion
with retained history~\cite{zhong2024memorybank,vermem}. These capacity- and
utility-driven forgetting mechanisms differ fundamentally from epistemic
\emph{temporal invalidation} during revision. Invalidation updates the temporal or contextual scope of a claim to record
that it no longer holds in the current state, without erasing the historical
evidence or evicting the record. Although both operations can reduce a
memory's future influence, factual validity (whether a claim still holds) and
retrieval availability (whether a record remains accessible) are distinct
properties. Choosing among these operations requires diagnosing whether the
issue is outdated knowledge, unwanted retrieval interference, or storage cost.

\paragraph{Selective Forgetting and Capability Preservation.}
Intentional forgetting aims to reduce unwanted influence, whereas
catastrophic forgetting refers to the unintended loss of knowledge or
capabilities that remain useful after further learning~\cite{luo2023catastrophic}. In external memory, useful information may also
become functionally unavailable even when the record survives~\cite{yu2026forget}. The presence of a \textsc{Delete} action in systems
such as AgeMem and UMA establishes a possible operation, not an effective
policy for selective forgetting~\cite{agemem,uma}. Nor does deleting a
source record necessarily remove its influence from summaries or procedures
derived from it. Conversely, consolidation can discard useful evidence
without any explicit deletion~\cite{zhang2026faultymemories,ramem}.
Where future usefulness is uncertain and continued retention is permitted,
reversible suppression or archival may be preferable to irreversible removal.
Maintenance should preserve still-useful knowledge, reduce harmful
influence, and retain enough evidence to recover from a premature update or
deletion. The next question is whether the policies that make these choices
can themselves improve from feedback.

\subsection{Feedback-Driven Policy Evolution}
\label{subsec:policy_evolution}

The preceding subsections describe how experience enters memory, how it is
organized and used, and how it is maintained as conditions change. We now
turn to the control of these operations, asking whether feedback can change
how the agent manages future experience.

\paragraph{Memory-State Adaptation versus Policy Evolution.}
The state and policy updates in Equations~\eqref{eq:memory_state_dynamics}
and~\eqref{eq:memory_policy_dynamics} target different objects of adaptation.
Persistent memory state $\mathcal{M}_t$ can continually evolve even when the
policy-update rule $\mathcal{U}$ leaves $\pi_t^{\mathrm{mem}}$ static, and a
fixed policy can already select different operations across contexts. To avoid
conflating these mechanisms, we distinguish three levels at which task
feedback can enter the lifecycle: (1) \emph{state-level incorporation}, where
feedback is written into $\mathcal{M}_t$ as new experience under a fixed
policy; (2) \emph{offline policy training}, where downstream task objectives
optimize $\pi^{\mathrm{mem}}$ prior to deployment (as in AgeMem and
UMA~\cite{agemem,uma}), after which the learned controller remains frozen;
and (3) \emph{online policy evolution}, where accumulated deployment feedback
continually modifies how future memory operations are selected or configured
($\pi_t^{\mathrm{mem}} \to \pi_{t+1}^{\mathrm{mem}}$). Online policy
evolution is therefore an advanced layer of adaptation rather than a
prerequisite for all memory-centric continual learning.

\paragraph{Feedback as a Control Signal.}
Existing systems illustrate how feedback operates across these levels. At the
state level, Reflexion stores verbal reflections derived from task outcomes,
and Voyager uses execution verification to admit procedures into its skill
library~\cite{shinn2023reflexion,wang2023voyager}; both improve subsequent
behavior by enriching $\mathcal{M}_t$ without altering the underlying memory
management rules. At the controller-training level, AgeMem and UMA optimize
memory-action selection against downstream rewards, while VerMem supplies
verifier-derived supervision for memory transitions during
training~\cite{agemem,uma,vermem}. The critical distinction lies in \emph{what}
the feedback updates and \emph{when} that update occurs: storing feedback
modifies available experience ($\mathcal{M}_t$), whereas training or adapting
a controller modifies the decision mechanism ($\pi^{\mathrm{mem}}$). Closing
the full policy-evolution loop requires keeping $\pi_t^{\mathrm{mem}}$ open
to continual revision during deployment across formation, consolidation,
retrieval, and maintenance, rather than treating feedback merely as a final
post-processing step.

\paragraph{Policy-Level Stability--Plasticity.}
When memory-management rules change, the stability--plasticity trade-off
also applies to the process that controls future learning. An overly
aggressive policy may repeatedly overwrite useful evidence, whereas an
overly conservative policy may reject experience needed for adaptation.
The effects of either policy can extend across many later memory
transitions. Learning from these effects is difficult because the value of
a write, consolidation, or deletion may become clear only in a later task.
A final task outcome may not reveal which earlier memory decision was
responsible. Verifier-based supervision can provide more local signals
during training, but it does not by itself establish the long-term value of
an operation~\cite{vermem}. Storage, latency, and computation costs must
also be weighed against future benefits. Repeated adaptation can erode
previously useful capabilities, which makes short-term gains an
insufficient guide to future updates~\cite{yu2026forget}. The objective is
therefore to improve memory management without repeatedly sacrificing
evidence or capabilities that remain useful.

\begin{figure}[t]
    \centering
    \includegraphics[
        width=\linewidth
    ]{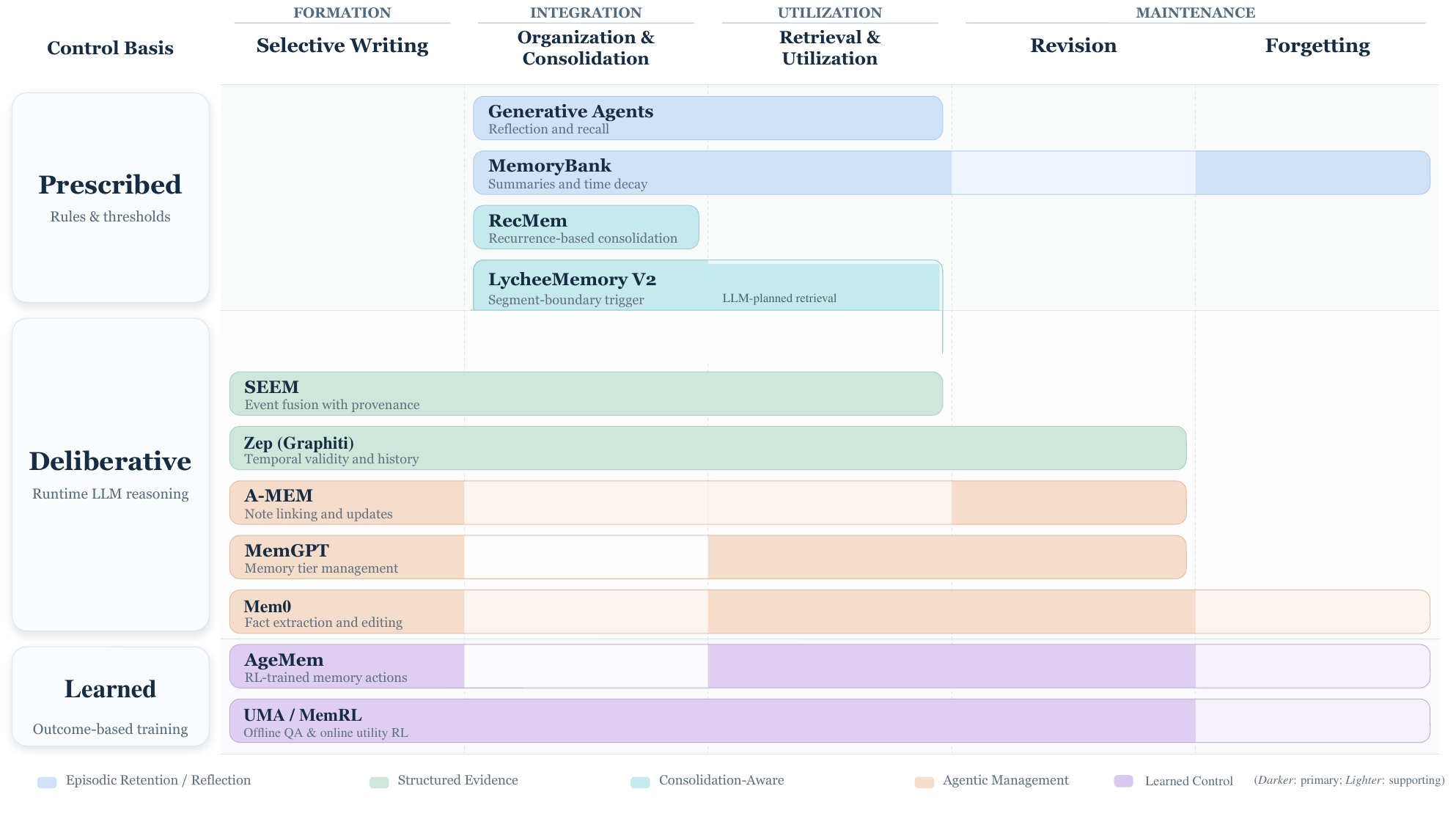}
    \caption{
        Lifecycle--control landscape of selected representative agent memory
        systems. The horizontal axis shows the memory operations emphasized
        by each system. The vertical axis shows how decisions are controlled:
        through prescribed rules, deliberative runtime LLM reasoning, or learned
        control. Darker regions indicate the main design focus and lighter
        regions indicate supporting operations. Colors distinguish design
        families. Positions describe design choices, not performance.
    }
    \label{fig:lifecycle_control_landscape}
\end{figure}

Figure~\ref{fig:lifecycle_control_landscape} brings the lifecycle operations
and their control mechanisms into a common view, mapping representative
systems to the operations they emphasize and the basis on which those
decisions are controlled. Across the horizontal axis, the figure groups
operations into formation (\emph{selective writing}), integration
(\emph{organization and consolidation}), utilization (\emph{retrieval and
utilization}), and maintenance, where \emph{revision} and \emph{forgetting}
are shown separately to distinguish the mechanisms discussed in
Section~\ref{subsec:revision_forgetting}. A single system may span several
operations and combine different forms of control.

The landscape shows overlapping design priorities rather than a single path
of development. Generative Agents and MemoryBank emphasize reflection,
summarization, and the selective reuse of past experience~\cite{park2023generative,zhong2024memorybank}. RecMem and LycheeMemory V2
make the timing and unit of consolidation explicit through semantic
recurrence and segment boundaries~\cite{recmem,li2026lycheememoryv2}.
SEEM, Zep (Graphiti), and A-MEM organize evolving memory through event fusion and
provenance, temporal fact management, and note linking and updates,
respectively~\cite{lu2026seem,rasmussen2025zep,xu2025amem}. MemGPT and Mem0
place selected decisions about memory tiers or records under runtime
deliberation~\cite{packer2023memgpt,chhikara2025mem0}. AgeMem, UMA, and MemRL train
memory decisions or retrieval utilities against downstream objectives~\cite{agemem,uma,zhang2026memrl}.
These emphases concern different parts of the learning process and can
coexist within a single system.

Two points guide the interpretation of this comparison. First, lifecycle
operation and control type are distinct design choices. LycheeMemory V2,
for example, combines prescribed segment-boundary rules with LLM-planned
retrieval~\cite{li2026lycheememoryv2}. A mechanism can respond to changing
inputs without its decision rule being learned from outcomes. Second,
similarly named operations may address different problems. Compressing
active context manages a short-term resource limit; the consolidation
discussed in Section~\ref{subsec:consolidation} organizes or transforms
experience for persistent reuse. Likewise, providing a \textsc{Delete}
operation permits removal but does not establish that the system chooses
what to forget appropriately. The figure should therefore be read as a
map of design choices, not a count of solved learning problems. The central
challenge is to coordinate these choices so that consolidation preserves
evidence needed for later use and maintenance reduces outdated influence
without damaging valid knowledge.

Taken together, the lifecycle explains how experience can change persistent
memory, how retrieval and utilization connect that memory to behavior, and how
revision and forgetting maintain it as evidence changes. Feedback may also
change the policies controlling these operations. Neither a memory update
nor a policy update, however, is sufficient evidence of successful
continual learning. The next section examines their behavioral consequences,
asking whether experience produces durable improvement, appropriate
adaptation, and transfer while preserving still-useful capabilities.

%% file: sections/evaluation.tex
\section{Evaluating Memory-Augmented Agents as Continual Learners}
\label{sec:evaluation}

Sections~\ref{sec:taxonomy} and~\ref{sec:memory_lifecycle} specify how
experience can persist, what reusable form it can take, and how it is
admitted, consolidated, used, revised, and governed. Those design choices do
not, by themselves, establish that continual learning has occurred. A
changed store, a retrieved record, or a higher single-task score can all
arise without a durable, experience-induced change in later behavior. This
section therefore treats evaluation as an evidentiary problem, asking what
empirical observations would justify a continual-learning claim given the
architecture and lifecycle established in prior sections. We first specify
three core behavioral outcomes (retention, adaptation, and transfer) and then
establish the protocols needed to measure them without confounding memory with
expanded context windows, extra test-time computation, or transient, isolated
success.

\subsection{What to Measure: Retention, Adaptation, and Transfer}
\label{sec:recall-not-learning}

Access-level metrics remain useful, but they answer a narrower question than
learning. Content correctness and retrieval scores diagnose whether
experience was written and can still be found; a single-task outcome shows
only that memory helped on the current probe
(Table~\ref{tab:recall-vs-learning}). Continual learning additionally
requires that an experiential gain persist under delay and interference,
remain revisable when the world changes, and transfer beyond remembered
episodes~\cite{lopezpaz2017gem,wang2024surveycl}. The remainder of this
subsection operationalizes those three outcomes. Component traces localize
failure; they do not substitute for the behavioral claim.

\begin{table}[t]
  \centering
  \caption{Diagnostic scope of evaluation metric layers in agent memory systems.
  While component-level and snapshot metrics diagnose internal storage fidelity and retrieval reachability,
  establishing continual learning requires behavioral retention, adaptive revision, and cross-task transfer.}
  \label{tab:recall-vs-learning}
  \small
  \renewcommand{\arraystretch}{1.2}
  \setlength{\tabcolsep}{5pt}
  \begin{tabular}{@{}
    >{\raggedright\arraybackslash}p{0.22\linewidth}
    >{\raggedright\arraybackslash}p{0.36\linewidth}
    >{\raggedright\arraybackslash}p{0.36\linewidth}
    @{}}
    \toprule
    \textbf{Metric Layer} & \textbf{Demonstrated Capability} & \textbf{Unaddressed Failure Mode} \\
    \midrule
    Content Fidelity
      & Accurate and lossless storage of past factual observations
      & Does not guarantee that stored evidence will be retrieved or utilized in future decisions \\
    Retrieval Access
      & High recall, precision, and reachability of relevant memories
      & Does not prevent retrieved evidence from being ignored, misinterpreted, or misapplied \\
    Single-Task Outcome
      & Immediate performance boost on a specific evaluation probe
      & Does not establish retention under delay and interference, adaptive revision, or transfer \\
    \bottomrule
  \end{tabular}
\end{table}

\subsubsection{Retention: Learning Must Survive Delay and Interference}
\label{sec:retention}

Section~\ref{subsec:retrieval_utilization} distinguishes preservation, access,
utilization, and behavioral effect in the memory lifecycle. Retention evaluation
uses the same cascade, but changes the object of measurement from a stored
trace to a capability margin. After delay and subsequent experience, the
agent should still outperform a matched agent deprived of that experience
on the same probe. An entry can remain in a store, or a latent state can
remain decodable, while its contribution to later decisions has vanished.
Conversely, a useful procedure may persist without the originating episode.
Trace survival is therefore diagnostic; behavioral persistence is the
learning claim.

The four gates yield a testable \emph{weakest-gate hypothesis}, positing that
behavioral retention is bottlenecked by the least reliable step and that
improving an earlier gate does not necessarily raise the final outcome.
Increasing capacity can raise preservation while lowering access through
denser retrieval competition. ``The system remembers $x$'' concerns preservation
and access; ``the system retained learning from $x$'' additionally requires
utilization and effect. With the base model fixed, continual insertion and
eviction in an external store can reduce performance on earlier
tasks~\cite{yu2026forget}, and repeated extraction and injection of experience
can turn a single-round gain into progressive
collapse~\cite{chen2026experienceinternalization}. These failures are not
confined to catastrophic forgetting in weights, as stored content can survive
while remaining unaccessed, unused, or ineffective.

Time and intervening experience are not interchangeable. A retention
protocol should first measure the immediate gain $\Delta_i^{0}$ on capability
$i$ relative to a memoryless or reset condition, then remeasure after delay $d$
and interference load $m$. The normalized retained gain
$\Delta_i(d,m)/\Delta_i^{0}$, reported only when $\Delta_i^{0}>0$, distinguishes
failure to retain from failure to learn. Delay should be specified in
wall-clock time, turns, or token distance; interference should report the number,
semantic overlap, and conflict relation of intervening updates.
Substrate-specific traces then localize the broken gate, identifying deletion,
eviction, and rank competition in external memory, or overwriting, state
drift, and readout mismatch in activation and parametric memory
(Section~\ref{subsec:2.1}). These outcomes are the quantities against which
capacity, interference, and cost are traded in
Section~\ref{sec:memory-scaling}.

\subsubsection{Adaptation: Learning Must Remain Revisable under Change}

When facts, preferences, tools, or rules change, retention without revision
becomes a source of error. Section~\ref{subsec:revision_forgetting} already
treats conflict resolution, persistent update, and selective forgetting as
lifecycle operations. Evaluation asks a more targeted question, examining whether
behavior faithfully follows a valid update after an environment change without
indiscriminate damage to still-useful capabilities. The relevant evaluation chain
is temporal rather than a single write (detection, revision, recovery, and
preservation), which should be quantified via detection delay, steps to
recovery, stale-error duration, and collateral degradation on unchanged probes.
Better recall does not automatically yield more
preference-aligned behavior~\cite{xiao2026alpsbench}, and summary memories
can retain factual detail while losing the preference signal needed after an
environmental shift~\cite{wang2026finperma}.

Component diagnostics should follow the substrate. For external stores, the
question is whether the current record supersedes the stale one and whether
retrieval ranks it above obsolete evidence. Exposed maintenance operations
establish that updates are possible, but conversational QA after a long
history does not demonstrate recovery under controlled
drift~\cite{chhikara2025mem0}. For parametric and latent memories, injection
accuracy and sequential fact updates are retention and write diagnostics,
not evidence that an acting agent detected an unannounced change or resumed
correct tool use~\cite{wang2024memoryllm,das2024larimar}. Evaluated models can
verbalize that a fact has changed yet fail to propagate that judgment to
the final prediction~\cite{wallat2026facts}. A credible adaptation claim
therefore requires matched pre-change and post-change probes, together with
simultaneous reporting of recovery and selective preservation.

\subsubsection{Transfer: Learning Must Extend beyond Remembered Episodes}

Section~\ref{sec:taxonomy} classifies reusable experience as episodic,
semantic, or procedural assets. Transfer evaluation asks whether those
assets support later behavior beyond storing an answer. Three increasingly
demanding cases should be separated: replaying the same or a near-duplicate
episode, solving an unseen variant from the same task family, and applying
shared structure to a distinct target task. Source and target partitions
must exclude duplicate instances and target answers, document the shared
structure, such as a workflow, causal rule, tool convention, or subskill, while
keeping the base model, tools, context budget, and compute matched. Otherwise
nearest-neighbour reproduction and extra context can be mistaken for
learned transfer.

Abstraction directly supports reuse, as demonstrated by the successful transfer
of insights, workflows, and executable skills across related tasks and
environments~\cite{zhao2024expel,wang2024awm,wang2023voyager}. The same
evidence shows why mean target accuracy is insufficient. Raw trajectory
memory can produce negative forward transfer where abstract insights do
not, with harm concentrated on cases that a no-memory agent could not
already solve~\cite{hu2026memorycl}. Dedicated memory and skill-maintenance
methods are only selectively helpful and can be matched or outperformed by
ordinary in-context accumulation~\cite{asawa2026clbench,guan2026continualskillbench,zhong2026skilllearnbench}.
Transfer should therefore be reported as zero-shot target gain or forward
transfer, interactions or tokens to a target threshold, and explicit
negative-transfer rates~\cite{lopezpaz2017gem,diazrodriguez2018metrics}.
For external memory, analysis should identify whether the retrieved asset
is a case, rule, workflow, or skill, and whether success depends on answer
overlap. For neural memory, it should test whether updated latent or
parametric state improves a held-out target without degrading unrelated
tasks.

\subsection{How to Measure: Protocols for Experience, Controls, and Time}

\subsubsection{Benchmarks Must Contain Learnable Experience}
\label{sec:learnable-experience}

A parallel distinction applies to benchmark design. The presence of
``history'' does not imply the presence of learnable experience. Many
suites pre-supply dialogue logs, profiles, or knowledge bases and then test
whether related questions can be answered after the history is
given~\cite{maharana2024locomo,du2024perltqa,wu2025longmemeval,zhong2024memorybank}.
Such protocols diagnose long-horizon access. They usually cannot decide
whether a new capability was formed \emph{during} evaluation.

Evidence strength rises with the experience that the protocol actually
exposes (Table~\ref{tab:benchmark-experience}). Fixed history supports ``remember and
use.'' Continual writing of state supports ``internal state accumulates.''
Feedback that improves later behavior, under the matched and sequential
controls below, approaches continual learning. Procedural streams ask
whether success and failure trajectories improve later
workflows~\cite{zheng2025lifelongagentbench}; semantic and personalization
streams ask whether facts and preferences persist and can be
revised~\cite{maharana2024locomo,wu2025longmemeval,tan2025membench}; dynamic-rule
streams ask whether invalidated experience is corrected in
time~\cite{hu2026memoryagentbench,asawa2026clbench}. A benchmark may excel
on one stream while remaining silent on the others. Externally grounded
outcomes and user corrections also support stronger learning claims than
self-generated reflections, which can be stored and later retrieved as if
they were facts~\cite{shinn2023reflexion,park2023generative}.

\begin{table}[t]
  \centering
  \caption{Categorization of representative agent memory benchmarks by evaluation paradigm.
  Benchmarks are organized by the extent to which their experimental design supports genuine
  learning-from-experience claims rather than static information retrieval.}
  \label{tab:benchmark-experience}
  \small
  \renewcommand{\arraystretch}{1.2}
  \setlength{\tabcolsep}{5pt}
  \begin{tabular}{@{}
    >{\raggedright\arraybackslash}p{0.30\linewidth}
    >{\raggedright\arraybackslash}p{0.28\linewidth}
    >{\raggedright\arraybackslash}p{0.36\linewidth}
    @{}}
    \toprule
    \textbf{Benchmark Family} & \textbf{Evaluation Paradigm} & \textbf{Supported Learning Claim} \\
    \midrule
    LoCoMo~\cite{maharana2024locomo}, LongMemEval~\cite{wu2025longmemeval}, PerLTQA~\cite{du2024perltqa}, MemoryBank~\cite{zhong2024memorybank}
      & Static dialogue history with single-turn question answering
      & Long-context factual recall, passive profile extraction, and needle retrieval \\
    MemBench~\cite{tan2025membench}
      & Multi-scenario synthetic memory evaluation
      & Access efficiency, query latency, and retrieval accuracy across storage scales \\
    MemoryAgentBench~\cite{hu2026memoryagentbench}
      & Incremental multi-turn ingestion with evolving facts
      & Dynamic knowledge assimilation, working-memory updates, and explicit conflict resolution \\
    LifelongAgentBench~\cite{zheng2025lifelongagentbench}, CL-Bench~\cite{asawa2026clbench}, AgentCL~\cite{shu2026agentcl}
      & Sequential interactive environments with task feedback
      & Cumulative experience accumulation, behavioral retention, and stateful margin ($\Delta P_t$) over reset baselines \\
    \bottomrule
  \end{tabular}
\end{table}

\subsubsection{Matched Comparisons Reveal What Memory Contributes}
\label{sec:matched_comparisons}

\begin{figure*}[t]
    \centering
    \includegraphics[width=0.72\textwidth]{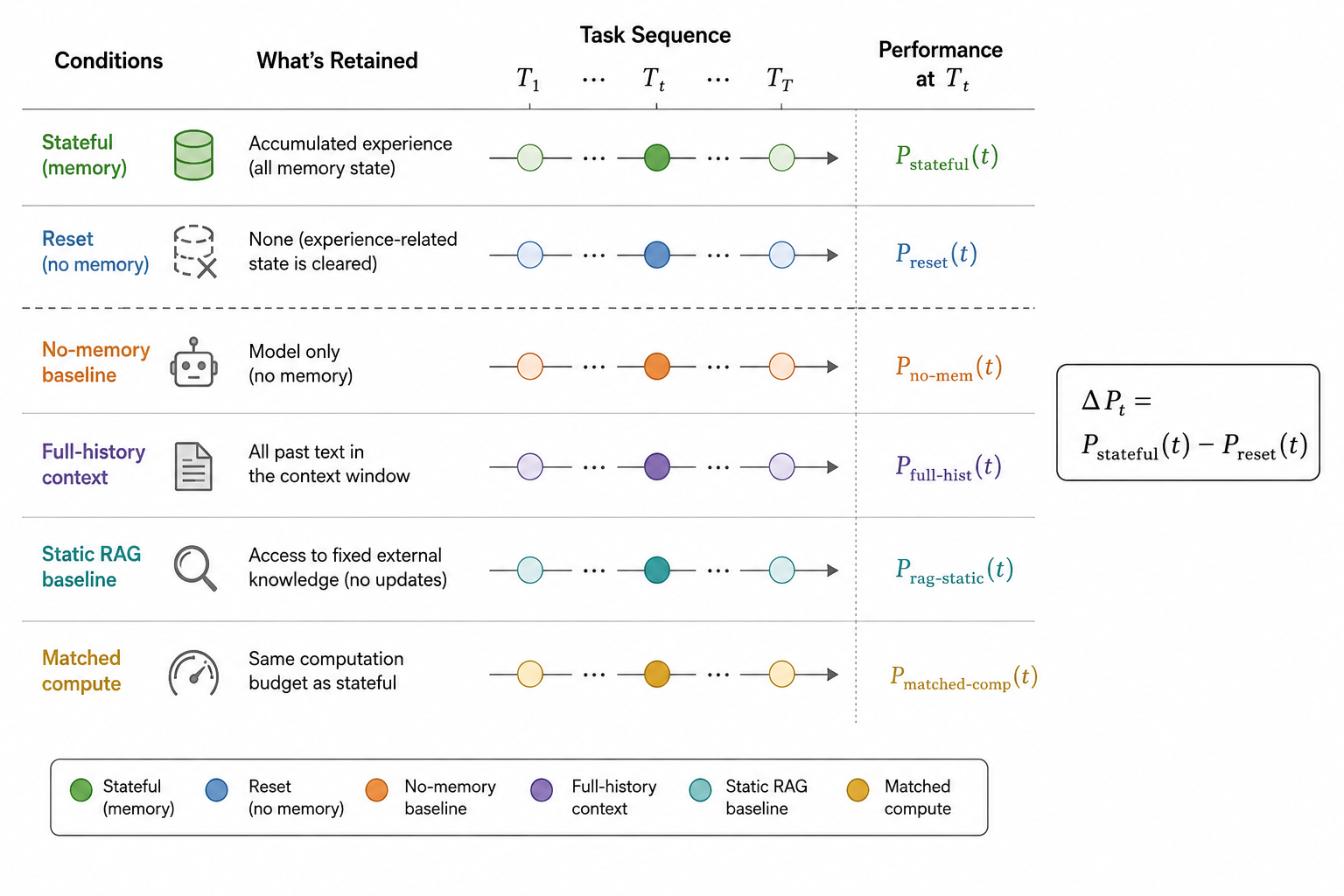}
    \caption{Matched comparison framework for isolating the contribution of
    agent memory. The stateful condition retains accumulated experience,
    whereas the reset condition removes experience-related state on a
    matched task. $\Delta P_t = P_{\mathrm{stateful}}(t) -
    P_{\mathrm{reset}}(t)$ is the additional benefit associated with
    accumulated state. No-memory, full-history, static RAG, and
    matched-compute baselines progressively rule out alternative
    explanations.}
    \label{fig:matched_comparison_framework}
\end{figure*}

Even when performance rises across a task sequence, the gain cannot be
attributed to memory until alternative explanations are removed. The
backbone, a longer context, static retrieval, or additional inference can
produce similar effects. Figure~\ref{fig:matched_comparison_framework}
isolates accumulated experience by comparing a stateful condition with a
reset condition on a matched task,
\begin{equation}
\Delta P_t
=
P_{\mathrm{stateful}}(t)
-
P_{\mathrm{reset}}(t).
\label{eq:memory_gain}
\end{equation}
$\Delta P_t$ is not a new composite metric. It denotes the additional
benefit associated with accumulated state under otherwise matched
conditions~\cite{tan2025membench,hu2026memoryagentbench,wei2025evomemory}.

Further baselines rule out further alternatives. A no-memory or reset
condition shows what the model can already do. A full-history context
baseline tests whether the benefit is simply more historical text; when a
memory mechanism outperforms a matched long-context condition, the evidence
favors organization or selective use rather than context expansion
alone~\cite{wang2026evomembench,tan2025membench}. A static RAG baseline
tests whether dynamic accumulation adds anything beyond a fixed knowledge
source. Matched computation addresses extra inference. Fine-tuning, LoRA,
and model editing absorb experience through different substrates and should
not be treated as interchangeable with all forms of agent memory, though
they remain useful comparison points.

What ``reset'' means depends on the substrate (Section~\ref{subsec:2.1}).
External stores admit write-off, read-off, and clear-store interventions,
which separate encoding, use, and overall value. For activation or
parametric memory, reset cannot simply mean deleting a database; evaluation must
instead distinguish online parameter updates, the preservation of persistent state, and
subsequent readout~\cite{gutierrez2025hipporag2}. A stateful condition may also
bundle conversation history, tool state, and external memory, so a positive
$\Delta P_t$ need not identify which mechanism produced the gain. A
stronger design keeps the backbone, task input, context budget, retrieval
budget, and inference compute matched while varying only
experience-related state.

\subsubsection{Sequential Evaluation Separates Durable Learning from Transient Success}
\label{sec:longitudinal-evidence}

The retention and adaptation definitions above already require the same
capability to be measured across memory states. Chronological length does
not substitute for that invariance. Endpoint scores after a long history
show that information can be used, but not when the benefit appeared or
whether it disappeared and recovered~\cite{maharana2024locomo,wu2025longmemeval}.
A short protocol that probes a capability before acquisition, immediately
afterward, and after a controlled update is stronger evidence than a
long trajectory with a single terminal score.

Four observation regimes support progressively stronger claims.
\emph{Snapshots} test a final memory state;
\emph{streams} report online reward on changing tasks;
\emph{reprobes} repeat the same capability or a calibrated task variant; and
\emph{interventional panels} combine reprobes with an event ledger and
matched stateful--reset trajectories. AgentCL and CL-Bench illustrate the
value of a stable observation target, where first-pass, frozen-memory, and
held-out probes systematically isolate acquisition, reuse, and
transfer~\cite{shu2026agentcl,asawa2026clbench}. Between probes, the ledger
should record elapsed time, writes, semantic overlap, conflicts, feedback,
tool or rule changes, and retrieval or update cost. Delay plus overlapping
updates tests retention; changes in facts, preferences, or tools test
adaptation; held-out tasks that preserve latent structure test transfer.
Evidence for one evaluation setting should not be treated as validation across
all three.

A compact blueprint follows. Define capability anchors and measure them
before and immediately after acquisition; schedule checkpoints on a declared
delay--interference grid; run stateful and reset conditions with identical
backbone, tools, context, retrieval, and inference budgets; and populate a
matrix $R_{t,i}$ that re-evaluates capability $i$ after update $t$.
Diagonal entries measure acquisition, lower-triangular entries expose
retention and backward influence, and held-out variants test transfer.
Reporting the event ledger beside this matrix prevents apparent gains
from obscuring task drift or hidden computational overhead.

\subsubsection{Fragmented Benchmarks Limit Cross-Study Conclusions}
\label{sec:fragmented-benchmarks}

Existing suites occupy different cells of this design rather than a common
scale. Access-oriented histories, incremental ingestion, and sequential
interactive tasks are not interchangeable measurements of the same
construct, and they rarely match context budgets, retrieval depth, or
inference compute. Numerical comparison across papers is therefore
unreliable unless the protocol specifies the behavioral outcome in
Section~\ref{sec:recall-not-learning}, the experience regime above, and the
matched and sequential controls that isolate memory. Scores should be
read within that evidentiary scope, not as a generic ranking of memory
systems.

\subsection{The Memory Scaling Law: Capacity, Interference, and Learning Efficiency}
\label{sec:memory-scaling}
\input{sections/scaling/nonparametric}
\input{sections/scaling/parametric}

%% file: sections/scaling/nonparametric.tex
\subsubsection{Non-Parametric Memory Scaling Law: Experience, Capacity, and Compute}
\label{sec:nonparametric-memory-scaling}

Non-parametric memory makes deployment experience a resource for continued
capability growth with fixed model weights. Agents preserve interaction
facts~\cite{chhikara2025mem0}, distill reasoning strategies, and consolidate
experience for later tasks~\cite{ouyang2025reasoningbank,yin2025experience}.
The resulting capability depends on how much relevant experience is
available and how efficiently the memory system retains and uses it.

We propose a Memory Scaling Law relating attainable performance to
experience, persistent capacity, and read--write compute. Its central
prediction is a sequence of resource-limited regimes, where performance
improves as relevant experience accumulates and subsequently saturates when
capacity or read--write compute constrains its effective use. Expanding the
limiting resources shifts these saturation points. The information structure of
experience determines how quickly capacity demand grows, while reuse determines
how much computation is worth investing in writing and consolidation. This view
connects memory representations and lifecycle operations to the rate and cost
of capability growth during deployment.

\paragraph{Resource dimensions and attainable performance.}
Let $E_t$ denote cumulative experience, $\bar B$ the persistent capacity
limit, $B_t$ actual occupancy, $R$ the per-query read budget, and $U$ the
write and maintenance budget over the same experience horizon. The budget
$U$ covers extraction, indexing, revision, and consolidation. These
quantities distinguish what an agent encounters, what it retains, and what
it can consult for a decision. Equal text volumes can contain different
amounts of task-relevant information, so a scaling relation also depends on
the source and composition of experience.

For a fixed foundation model $\theta$, task distribution $\mathcal D$,
experience-generation process, and remaining inference settings, let
$\epsilon^\star(E,\bar B,R,U\mid\theta,\mathcal D)$ be the minimum expected
task error within a specified class of memory policies. Task error is an
error rate or a gap in normalized task score. This attainable frontier
separates resource limits from the performance of an individual method.
Capacity accounts for all persistent experience-dependent state, including
informative vectors and indices; text-token occupancy measures its textual
part.

\paragraph{Joint scaling and resource-optimal allocation.}
Within a regime of relatively stable experience quality and task structure,
we hypothesize
\begin{equation}
 \epsilon^\star-\epsilon_\infty
 \asymp \max_{j\in\mathcal A} a_j x_j^{-\alpha_j},
 \qquad
 \mathbf x=\left(\frac{E}{E_0},\frac{\bar B}{B_0},
                  \frac{R}{R_0},\frac{U}{U_0}\right),
 \quad a_j,\alpha_j>0.
 \label{eq:ms-joint-scaling}
\end{equation}
The model and task family determine the limiting error $\epsilon_\infty$;
$\epsilon_{\mathrm{ex}}=\epsilon^\star-\epsilon_\infty$ denotes excess task error.
Each dimensionless $x_j$ measures scale relative to a reference budget, and
$\mathcal A$ contains the active resource constraints. The largest term
approximates the error from the dominant shortage; an additive form gives a
smooth approximation when several terms are comparable. The exponents and
the ranges over which they apply are empirical quantities.

A long-tailed demand distribution gives a simple capacity example. Suppose
each query requires one knowledge unit, unit demand satisfies
$p_i\propto i^{-s}$ with $s>1$, and each unit occupies $\tau$ tokens. If the
experience pool contains all units and the most requested $K=\lfloor\bar B/\tau\rfloor$
units are stored and reliably accessed, uncovered demand is
$\epsilon_{\mathrm{cov}}(\bar B)=\sum_{i>K}p_i\propto\bar B^{-(s-1)}$.
Then $\alpha_B=s-1$: the capacity exponent follows from the distribution of
task-relevant demand under these assumptions.

\begin{figure}[!htbp]
 \centering
 \includegraphics[width=.90\linewidth]{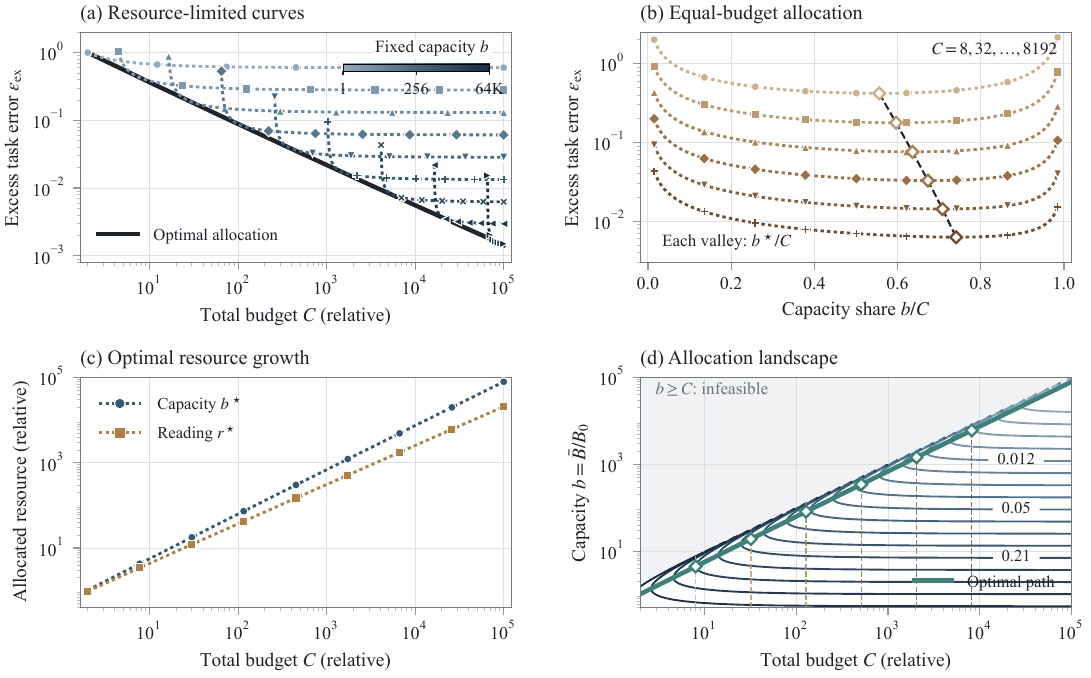}
 \caption{Theoretical illustration of joint scaling and optimal resource allocation. (a) Fixed-capacity error curves and their optimized lower envelope. (b) Returns at equal total budgets. (c) Optimal capacity and reading budgets. (d) Equal-error contours and the optimal path. For normalized capacity $b=\bar B/B_0$ and reading $r=R/R_0$, curves use $\epsilon_{\mathrm{ex}}(b,r)=0.6b^{-0.55}+0.4r^{-0.75}$ and $C=b+r$, with other resources non-binding. Filled markers sample this function; hollow diamonds mark its calculated optima.}
 \label{fig:ms-law}
\end{figure}

A capacity-only sweep can plateau because reading, experience supply, or
maintenance becomes limiting. Expanding the limiting resource moves the
plateau, allowing further gains from capacity (Figure~\ref{fig:ms-law}).
The joint relation thus predicts both diminishing returns within a fixed
configuration and renewed improvement when resource allocation changes.
Keeping the active experience, capacity, and reading terms at comparable
orders gives a balanced scaling path:
\begin{equation}
 \frac{\bar B}{B_0}\propto
 \left(\frac{E}{E_0}\right)^{\alpha_E/\alpha_B},
 \qquad
 \frac{R}{R_0}\propto
 \left(\frac{E}{E_0}\right)^{\alpha_E/\alpha_R}.
 \label{eq:ms-balanced-scaling}
\end{equation}
Maintenance follows an analogous relation. Resources need not grow in equal
proportions. Under a lifecycle budget $C$ and $N$ reuses, the cost function
$\mathcal C_N$ also determines the optimal allocation:
\begin{equation}
 \mathbf x^\star(C,N)\in
 \operatorname*{arg\,min}_{\mathbf x:\,\mathcal C_N(\mathbf x)\leq C}
 \epsilon^\star(\mathbf x\mid\theta,\mathcal D).
 \label{eq:ms-budget-optimum}
\end{equation}
The frontier is specific to the policy class: admitting more efficient
representations or access mechanisms can change its shape and its optimal
allocation. Foundation-model capabilities also condition the returns.
Pretrained knowledge reduces some storage requirements, while the ability to
interpret abstractions and combine evidence increases the value of a given
memory bank. Recent facts and private interactions have to be acquired
during deployment.

The measured comparisons in Table~\ref{tab:ms-protocol} use
\texttt{Qwen3-30B-A3B-Instruct-2507} for construction, answering, and local
model-based scoring, with 1,024-dimensional \texttt{qwen3-embedding-0.6b}
embeddings. We evaluate Mem0 0.1.94~\cite{chhikara2025mem0} and retrieval over
retained raw histories (Raw RAG).
Capacity sweeps use frozen bank subsets; continual tasks update memory under
a capacity limit. Stored and read text is counted with \texttt{o200k\_base}.
Scores and occupancy are macro-averaged over conversations or samples, with
cluster-bootstrap $95\%$ intervals. LongMemEval uses a local semantic judge,
and MemoryBench uses local rubric scoring.

\begin{figure}[!htbp]
 \centering
 \includegraphics[width=.94\linewidth]{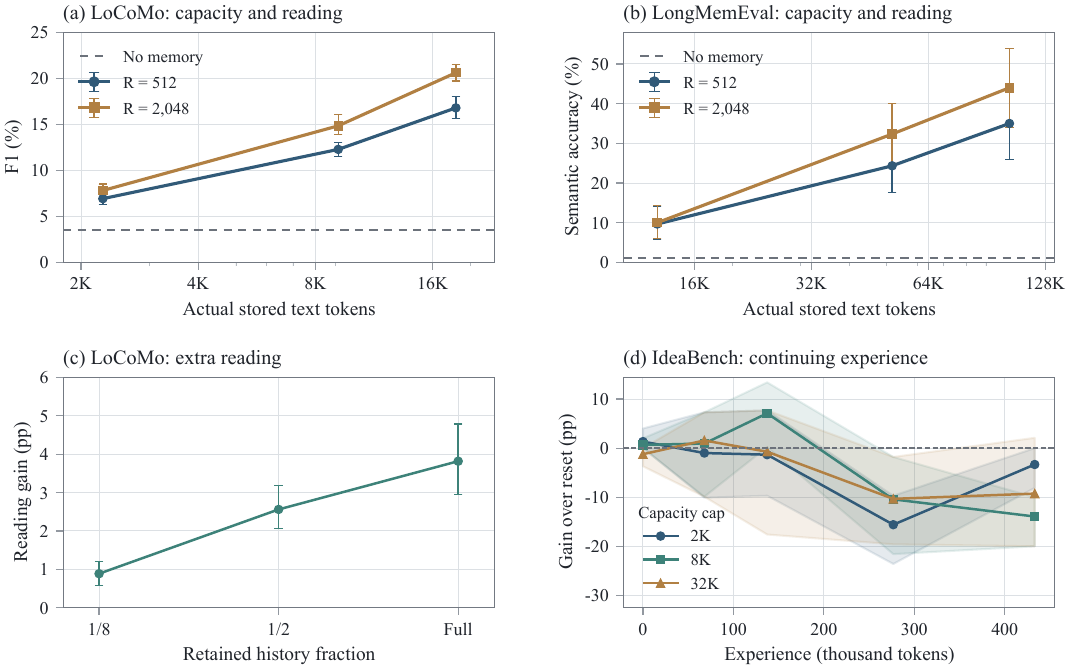}
 \caption{Measured performance across capacity, reading, and experience settings. (a--b) Stored text and Raw RAG performance on LoCoMo and LongMemEval. (c) Paired gains from increasing LoCoMo reading from $512$ to $2{,}048$ tokens; error bars in (a--c) are cluster-bootstrap $95\%$ intervals. (d) Stateful--reset score differences on fixed IdeaBench questions at successive checkpoints, without periodic consolidation; shading spans the min--max range across three random training-dialogue orderings. Continuous horizontal axes report text-token volumes; (c) indexes retained-history fractions, and capacity labels in (d) denote upper limits.}
 \label{fig:ms-storage-read}
\end{figure}

While the additive approximation in Figure~\ref{fig:ms-law} predicts that
storage and reading co-scale under a shared budget constraint, empirical
history-based question answering also exhibits direct interaction because
effective readout is bounded by retained storage ($R_{\mathrm{eff}}=\min(R,B_t)$)
and retrieval competition (Figure~\ref{fig:ms-storage-read}(a--c)). On LoCoMo~\cite{maharana2024locomo},
increasing the read budget from 512 to 2,048 tokens improves Raw RAG F1 by
0.89 percentage points at one-eighth history retention and 3.82 points at
full retention. The paired difference between these gains is 2.93 points
($95\%$ interval: $[2.06,4.02]$). On the selected
LongMemEval~\cite{wu2025longmemeval} samples, full-history Raw RAG accuracy
increases from 35.00\% to 44.00\% over the same read-budget change.
These observations confirm storage--reading complementarity: the return
from additional reading depends on the available memory.

\begin{figure}[!htbp]
 \centering
 \includegraphics[width=.96\linewidth]{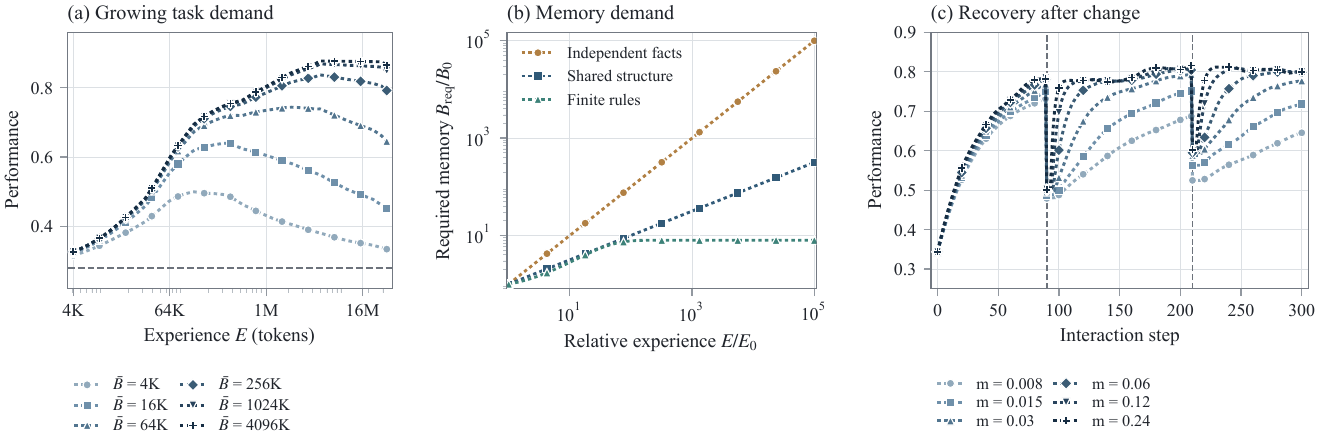}
 \caption{Theoretical illustration of experience structure, capacity requirements, and adaptation. Curves use preset response and recovery functions. (a) Performance as task demand expands with experience, at fixed capacity limits $\bar B$. (b) Capacity needed to maintain a target under different experience structures. (c) Recovery after environmental changes, marked by dashed lines; $m$ is the maintenance rate.}
 \label{fig:ms-experience}
\end{figure}

\paragraph{Experience structure and capacity requirements.}
During deployment, experience can introduce new task demands, so the relevant
distribution evolves as $\mathcal D_E$. At fixed capacity, early gains can
give way to decline when the information needed to serve this expanding
workload outgrows the bank (Figure~\ref{fig:ms-experience}(a)). For $n$
independent, equiprobable binary facts that are independent of the fixed
model and queried uniformly, keeping average reconstruction error below
$\delta<1/2$ requires at least $n[1-h_2(\delta)]$ bits, where $h_2$ is binary
entropy. Shared rules and skills permit more compact representations; a
finite set of recurring task structures can produce saturating demand.
\begin{table}[!htbp]
\centering
\caption{Experimental comparisons in this section. Repeated predictions across budgets are aggregated within the stated sampling units. LoCoMo scores use categories 1--4.}
\label{tab:ms-protocol}
\small
\renewcommand{\arraystretch}{1.2}
\setlength{\tabcolsep}{5pt}
\begin{tabularx}{\linewidth}{@{}l>{\raggedright\arraybackslash}X>{\raggedright\arraybackslash}X@{}}
\toprule
\textbf{Empirical Study} & \textbf{Experimental Protocol \& Controls} & \textbf{Measured Outcomes} \\
\midrule
LoCoMo Capacity Sweep & 10 long-form dialogues, 1,540 questions; $4 \times 4$ storage--read capacity grid & Lexical F1 score with cluster-bootstrap 95\% CI \\
LongMemEval Capacity Sweep & 100 selected samples; storage--read capacity grid & Local semantic accuracy \\
IdeaBench Sequential Deployment & 50 fixed target queries evaluated across sequential training checkpoints (3 random task orderings) & Stateful--reset performance margin ($\Delta P_t$) without consolidation \\
LoCoMo Query Economics & Cohort of 9 conversations, 180 questions; comparing direct read, search, and search + summarize & F1 score, serialized context tokens, and end-to-end query latency \\
\bottomrule
\end{tabularx}
\end{table}

Let $B_{\mathrm{req}}(E;p_0)$ be the minimum capacity needed to maintain
performance $p_0$ as relevant task demand expands with experience. With
adequate reading and maintenance, we describe its growth within a stable
regime as
\begin{equation}
 B_{\mathrm{req}}(E;p_0)\propto E^{\nu}.
 \label{eq:ms-memory-demand}
\end{equation}
The exponent $\nu$ describes capacity demand, while $\alpha_j$ describes the
performance return from a resource. Independent information, shared
structure, and recurring rules yield different demand curves
(Figure~\ref{fig:ms-experience}(b)). A system that preserves separate episodes
and one that extracts reusable rules can therefore require different
capacity growth to sustain the same target performance.

Consolidation changes this relationship by turning repeated detail into
reusable structure. ReasoningBank converts successes and failures into
reasoning strategies~\cite{ouyang2025reasoningbank}. Experience Scaling merges
and refines experience to reduce redundancy, while finding that excessive
compression can harm performance~\cite{yin2025experience}. Abstractions can
support multiple tasks, while exceptions and rare queries may need the
original details. Episodic, semantic, and procedural assets thus have
different fidelity and capacity requirements. Under environmental change,
the rate of revision also determines how quickly useful performance can
recover (Figure~\ref{fig:ms-experience}(c)).

\FloatBarrier

Across the evaluated IdeaBench checkpoints from MemoryBench~\cite{ai2025memorybench},
Mem0's stateful--reset score difference varies non-monotonically
(Figure~\ref{fig:ms-storage-read}(d); Mem0 is run in its default online mode, which also writes during the 50-query evaluation pass and produces small non-zero offsets at $t=0$). At an 8,192-token capacity limit and
$R=1{,}024$, mean stateful--reset gain falls from $+7.10$ percentage points after
64 training dialogues to $-10.43$ after 128 and $-13.93$ at the final 256-dialogue checkpoint (because unconstrained occupancy plateaus at 3,858 tokens, neither the 8,192- nor the 32,768-token cap binds; pooling both 3-ordering replicates gives $+3.18$ points at 64 dialogues, $-10.38$ at 128, and $-11.60$ at 256).
Consolidation every 16 or 64 training dialogues reduces final occupancy from
3,858 to approximately 1,800 tokens, while final gains remain negative.

% Keep the short final evidence table from stretching paragraph spacing.
\raggedbottom
\paragraph{Selective access and the economics of reuse.}
Selective access allows persistent memory and per-query input to scale
separately. If a query needs a bounded number of fixed-length records and
indexing reliably locates them, model input can stay bounded as the bank
grows. Tasks that synthesize more evidence may require more reading, and
index search and updates still incur costs. Cross-task reuse can shift work
from repeated reading and reasoning to earlier extraction or consolidation.
With costs expressed in consistent units, average cost over $N$ reuses is
\begin{equation}
 \overline C_N=\frac{C_{\mathrm{build}}+C_{\mathrm{maint}}(N)}{N}
                +\overline C_{\mathrm{query}}.
 \label{eq:ms-lifecycle-cost}
\end{equation}
At equal performance, lightweight writing and direct reading may cost less
when reuse is limited. More expensive preparation becomes worthwhile when
cumulative query savings exceed its additional cost. For fixed-quality
policies with no ongoing maintenance, the crossover reuse count is the
extra construction cost divided by the per-query saving.

\begin{figure}[!htbp]
 \centering
 \includegraphics[width=\linewidth]{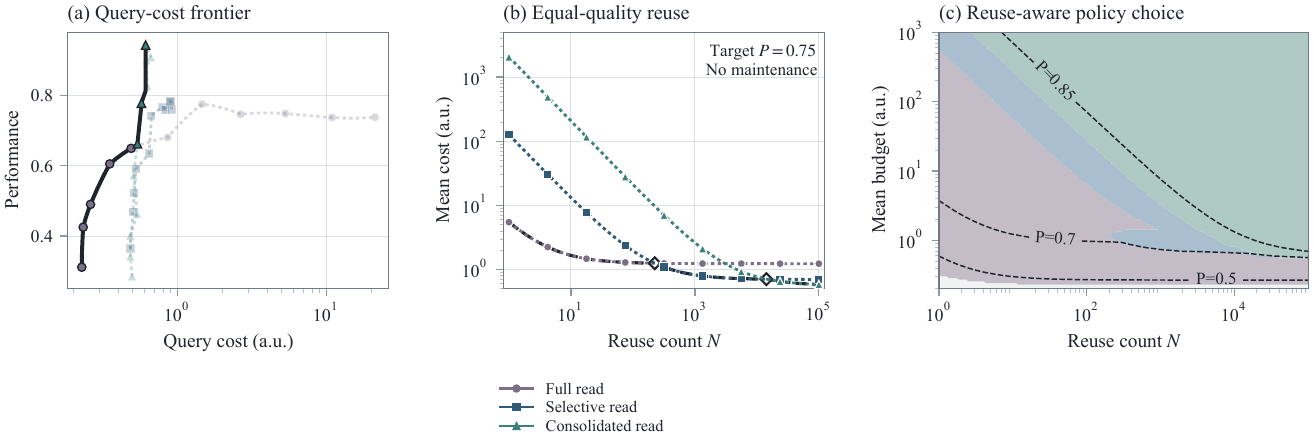}
 \caption{Theoretical illustration of reuse and resource-optimal policy choice. (a) Query-cost--performance Pareto frontiers; lighter points denote dominated configurations. (b) Amortization at a common performance target $P=0.75$. (c) The best policy at each reuse count and average-cost budget; gray denotes infeasibility, and dashed lines are performance contours. Preset functions specify policy performance and costs in common arbitrary units, with offline consolidation and no ongoing maintenance.}
 \label{fig:ms-policy-frontiers}
\end{figure}

Token savings and latency savings can differ. On the matched LoCoMo
small-bank cohort (Table~\ref{tab:ms-query-cost}), query-time summarization
reduces serialized memory input but increases latency. Its paired F1 is
$0.72$ percentage points below full-bank reading ($95\%$ interval for the
difference: $[-1.14,-0.27]$). Policy choice consequently depends on which
resource is constrained and the performance target.

\begin{table}[!htbp]
\centering
\caption{Three reading policies over Mem0 memory banks on nine LoCoMo conversations and 180 questions, with three timing repetitions. Memory input includes serialization and formatting. Median latency covers retrieval, query-time summarization, and answering, excluding offline construction. F1 intervals use $10{,}000$ conversation-cluster bootstrap resamples.}
\label{tab:ms-query-cost}
\small
\setlength{\tabcolsep}{5pt}
\begin{tabular}{@{}lrrrr@{}}
\toprule
Reader & F1 (\%) $\uparrow$ & 95\% interval & Input (tokens) & Latency (s) $\downarrow$ \\
\midrule
Read all & 4.63 & [3.28, 6.02] & 729.78 & 0.154 \\
Search & 4.29 & [3.15, 5.50] & 225.68 & 0.154 \\
Search + summarize & 3.91 & [2.81, 5.15] & 43.52 & 0.367 \\
\bottomrule
\end{tabular}
\end{table}

Environmental change bounds how many times a memory can be reused before
it needs revision. Retention extends the period over which preparation costs
can be amortized, and transfer increases the number of tasks served by the
same experience. Adaptation incurs the maintenance cost required to keep
that experience useful. These continual-learning properties determine the
reuse horizon and update costs in the resource-optimal allocation of
Equation~\eqref{eq:ms-budget-optimum}.
\FloatBarrier
\ifdefined\afterpage
  \afterpage{\begingroup\globaldefs=1\flushbottom\endgroup}
\else\flushbottom\fi

%% file: sections/scaling/parametric.tex
\subsubsection{Parametric Memory Scaling Law}
\label{sec:parametric-memory-scaling}

Parametric memory writes deployment experience into model parameters or
persistent neural states rather than into an external store. Representative
designs include trainable memory layers~\cite{berges2024memorylayers},
plug-and-play memory modules~\cite{cao2025memorydecoder}, memory states
updated at test time~\cite{behrouz2025titans,kuratov2026gradmem}, side
memories isolated from the backbone~\cite{wang2024wise,das2024larimar,wang2024memoryllm},
and direct editing or fine-tuning of backbone 
weights~\cite{meng2023memit,xu2026loraremembers}. These approaches confront the
same fundamental question as non-parametric memory regarding how performance
scales with experience and resources, and at what write and compute cost. They
differ in one essential respect. All writes share the same parameters, so new
writes alter how earlier content is read out. Accumulating experience therefore
adds usable content while simultaneously increasing the probability that
existing content is overwritten. We build this property into the scaling relation
and propose a parametric counterpart to Equation~\eqref{eq:ms-joint-scaling}.

\paragraph{Resource dimensions of parametric memory.}
We retain the notation of Section~\ref{sec:nonparametric-memory-scaling},
where $E_t$ denotes cumulative experience, $\theta$ denotes the foundation model,
and $\mathcal D$ denotes the task distribution. Parametric memory involves three
resources. Memory capacity $M$ is the number of memory parameters, memory
slots, adapter rank, or persistent-state dimensions, playing the role of
$\bar B$. The write budget $U$ represents the computation used to write
experience into parameters, such as gradient steps, update steps per write, and
write frequency. Readout compute is fixed by the architecture, as each forward
pass traverses the memory parameters or performs a fixed top-$k$ lookup
without consuming additional prompt context tokens;
consequently, the read budget $R$ is generally not an independent decision
variable. This marks the first major divergence from non-parametric memory. 

The second difference concerns occupancy. External memory can directly report its
actual occupancy $B_t$, whereas parametric memory has no directly observable
counterpart. We therefore define the load ratio
\begin{equation}
 \rho_t=\frac{I(E_t)}{\kappa M}, 
 \label{eq:pm-load}
\end{equation}
where $I(E_t)$ is the information in the experience that must be retained,
measured in bits, and $\kappa$ is the information a single parameter can
store. Knowledge-capacity studies estimate $\kappa\approx 2$ bits per
parameter and report that this ratio holds even under int8
quantization~\cite{allenzhu2024capacity}. The dimensionless quantity
$\rho_t$ measures how full the memory has become.

\paragraph{Joint scaling with an interference term.}
Within a regime of relatively stable experience quality and task structure,
we hypothesize
\begin{equation}
 \epsilon^\star-\epsilon_\infty
 \asymp \max_{j\in\mathcal A} a_j x_j^{-\alpha_j}
 + I_{s,\pi}(\rho_t),
 \qquad
 \mathbf x=\left(\frac{E}{E_0},\frac{M}{M_0},\frac{U}{U_0}\right).
 \label{eq:pm-joint-scaling}
\end{equation}
The first term mirrors Equation~\eqref{eq:ms-joint-scaling}, approximating the
residual loss from resource shortages where the largest component identifies the
active bottleneck. The
second term is an interference term specific to parametric memory. It is
close to zero while the load ratio stays below a critical value
$\rho^\star_{s,\pi}$; beyond that value, new writes begin to overwrite
earlier content and loss rises. The critical ratio depends on the memory
substrate $s$ (Section~\ref{subsec:2.1}) and on the write policy $\pi$. In
non-parametric memory, additional experience can crowd retrieval or trigger
eviction while leaving surviving records intact, a cost mediated by the read
and maintenance terms. In parametric
memory, additional experience rewrites shared parameters directly, so performance
degradation as experience grows is an intrinsic property of the substrate
rather than only a retrieval or selection problem.

\begin{figure}[!htb]
 \centering
 \includegraphics[width=.96\linewidth]{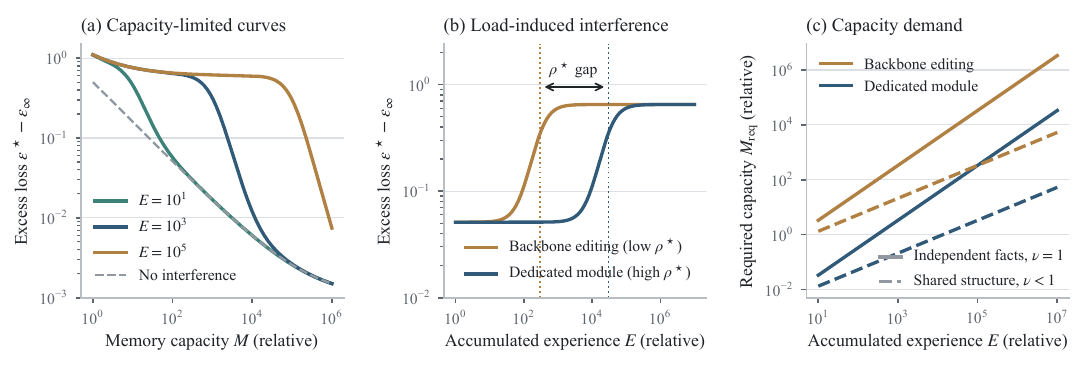}
 \caption{Theoretical illustration of the parametric memory scaling relation. (a) Excess loss versus memory capacity at different experience volumes; interference dominates until the load ratio falls below $\rho^\star$, after which the capacity term follows a power law. (b) Excess loss versus accumulated experience at fixed capacity for backbone editing (low $\rho^\star$) and a dedicated memory module (high $\rho^\star$); dotted lines mark the critical loads. (c) Capacity required to maintain a target performance for independent facts ($\nu=1$) and shared structure ($\nu<1$). Curves sample the illustrative instance of Equations~\eqref{eq:pm-load}--\eqref{eq:pm-memory-demand} described in the text.}
 \label{fig:pm-law}
\end{figure}

Equation~\eqref{eq:pm-joint-scaling} yields three testable predictions.
First, when experience and write budget are fixed and only capacity grows,
loss is initially dominated by interference and follows the power law only
after the load ratio enters the safe regime (Figure~\ref{fig:pm-law}(a)).
Second, when capacity is fixed and experience grows, loss rises once the
load ratio crosses $\rho^\star$, and the transition can differ by orders of
magnitude across substrates (Figure~\ref{fig:pm-law}(b)). Third, load curves
measured at several capacities should approximately collapse onto a single
curve when plotted against $\rho$. This collapse is the most direct test of
whether interference depends on load only through the ratio.

Figure~\ref{fig:pm-law} illustrates these predictions with a minimal
instance of Equation~\eqref{eq:pm-joint-scaling}. It keeps only the capacity
term and the interference term,
$\epsilon^\star-\epsilon_\infty=a_M M^{-\alpha_M}+I(\rho)$, with
$I(\rho)=\lambda(\rho/\rho^\star)^{b}/[1+(\rho/\rho^\star)^{b}]$ and
$\rho=E/M$ in relative units ($\kappa=1$). We set $a_M=\alpha_M=0.5$,
$\lambda=0.6$, and $b=2$. Backbone editing and a dedicated module differ
only in the critical ratio, $\rho^\star=3$ and $\rho^\star=300$, while
panel~(a) uses $\rho^\star=1$. Panel~(c) plots
$M_{\mathrm{req}}=E^{\nu}/\rho^\star$ from Equation~\eqref{eq:pm-memory-demand}
with $\nu\in\{1,0.6\}$.

\begin{figure}[!htb]
 \centering
 \includegraphics[width=.94\linewidth]{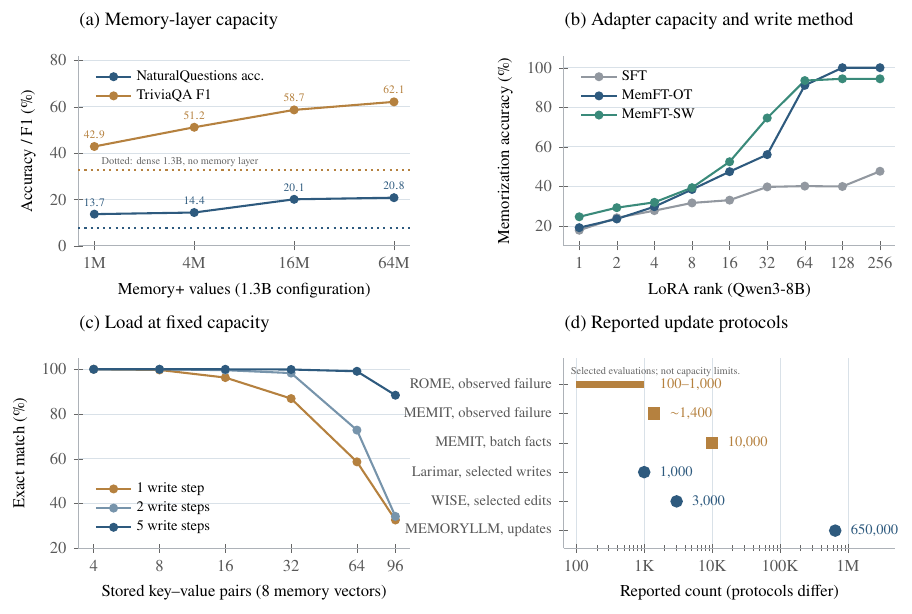}
 \caption{Reported capacity and load effects in published parametric-memory studies. (a) Memory+ value count at a fixed 1.3B configuration~\cite{berges2024memorylayers}; dotted lines show the dense model without memory layers. (b) Memorization accuracy versus LoRA rank for three write methods on Qwen3-8B~\cite{xu2026loraremembers}. (c) Reported exact match versus stored key--value pairs with eight fixed memory vectors and different numbers of write steps~\cite{kuratov2026gradmem}. (d) Reported sequential or batch update horizons across parametric-memory studies~\cite{meng2022rome,meng2023memit,gupta2024editingscale,das2024larimar,wang2024wise,wang2024memoryllm}; MEMORYLLM evaluates recent-context retention across 650,000 context updates under random slot replacement rather than simultaneous retention of all past edits. All values are taken from the original papers without fitting.}
 \label{fig:pm-evidence}
\end{figure}

\paragraph{Evidence for the capacity term.}
Experiments that hold the foundation model fixed and enlarge only the memory
parameters support the shape of the capacity term. Memory Layers at Scale
enlarges the Memory+ value count from 1M to 64M in a 1.3B configuration
(Figure~\ref{fig:pm-evidence}(a))~\cite{berges2024memorylayers}.
NaturalQuestions accuracy rises from 13.68 to 20.78 and TriviaQA F1 from
42.89 to 62.14, well above a dense model of the same configuration without
memory layers (7.76 and 32.64). Between 16M and 64M, however,
NaturalQuestions improves by only 0.64 while TriviaQA still improves by 3.47,
indicating that the same capacity range places different tasks at different
points on the power law. The same work scales memory parameters to 128B on
foundation models of at most 8B, showing that memory capacity can far exceed
backbone size (though these Memory+ layers are trained jointly during pretraining,
so sequential deployment writes remain to be tested separately).

Direct functional evidence comes from LoRA memorization. Treating
LoRA rank $r$ as a capacity probe, a recent study reports a power law
$\Delta\mathcal L=C\,r^{\alpha}\ell^{-\beta}+b$ that links the loss
reduction to rank and sequence length $\ell$ across models, datasets, and
semantic densities ($R^2>0.98$)~\cite{xu2026loraremembers}. The relation
contains both a capacity variable and a load variable. If
$\alpha\approx\beta$, it reduces to a function of $r/\ell$, the dependence
on the load ratio that Equation~\eqref{eq:pm-joint-scaling} predicts;
whether the two exponents coincide is therefore a direct test of the
proposed form. The same study shows that capacity alone does not determine
what is stored. At equal rank on Qwen3-8B, memorization accuracy differs
sharply across write methods (Figure~\ref{fig:pm-evidence}(b)), where at rank 128,
SFT reaches $40.0\%$, MemFT-OT reaches $100.0\%$, and MemFT-SW reaches $94.4\%$.
In the terms of Equation~\eqref{eq:pm-joint-scaling}, the write method changes
$a_M$ and $\alpha_M$, so better methods lie closer to the attainable frontier at
the same capacity.

\begin{figure}[!htb]
 \centering
 \includegraphics[width=.96\linewidth]{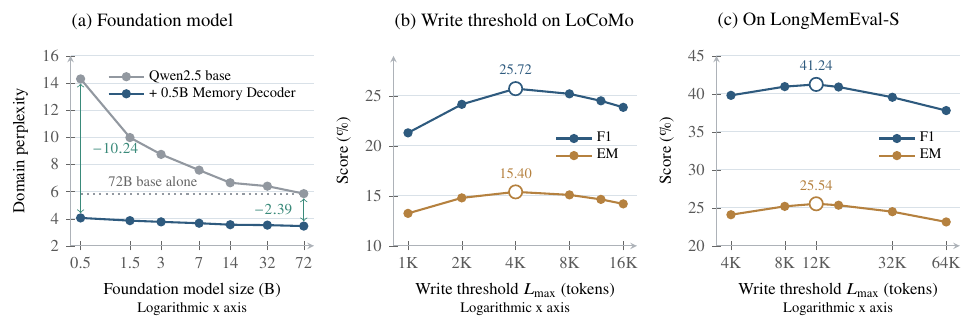}
 \caption{Reported effects of the foundation model and write timing on parametric memory. (a) Average domain perplexity (biomedicine, finance, and law) of Qwen2.5 models from 0.5B to 72B, with and without the same 0.5B Memory Decoder~\cite{cao2025memorydecoder}; arrows mark the reduction at the smallest and largest models, and the dotted line marks the 72B model alone. (b--c) F1 and exact match of TMEM on Qwen3-4B as the context threshold that triggers a parametric write varies, on LoCoMo and LongMemEval-S~\cite{ren2026tmem}; hollow markers denote the observed maxima. A larger threshold means less frequent writes, not more write compute. All values are taken from the original papers without fitting.}
 \label{fig:pm-conditions}
\end{figure}

The foundation model also conditions memory returns. Attaching the same 0.5B
Memory Decoder to Qwen2.5 models from 0.5B to 72B reduces average domain
perplexity by 10.24 at 0.5B but by only 2.39 at 72B, and the 0.5B model with
the memory module (4.06) outperforms the 72B model alone
(5.85; Figure~\ref{fig:pm-conditions}(a))~\cite{cao2025memorydecoder}. This mirrors the substitution and
complementarity between model and memory discussed in
Section~\ref{sec:nonparametric-memory-scaling}. Continual fine-tuning studies
further find that, between 1B and 7B parameters, larger models forget more
severely~\cite{luo2023catastrophic}. The foundation model may therefore shift
$\rho^\star$ as well as $\epsilon_\infty$ and should be reported as a
controlled condition.

\paragraph{Evidence for the load term and the critical ratio.}
Figure~\ref{fig:pm-evidence}(d) summarizes the number of updates or evaluated
write horizons reported across representative parametric-memory studies. With
direct backbone editing, sequential ROME edits degrade performance
after roughly 100 to 1,000 edits across typical edit sequences and can even trigger immediate model collapse after certain individual disabling edits~\cite{meng2022rome,gupta2024editingscale}. Sequential MEMIT enters
catastrophic forgetting after about 1,400 edits on GPT-J 6B, whereas batch
MEMIT on the same model edits 10,000 facts simultaneously in a single batch~\cite{gupta2024editingscale,meng2023memit}. When writes are isolated
from the backbone, Larimar sustains 1,000 sequential scope writes (200 facts across 5 rephrasings)~\cite{das2024larimar}, WISE remains effective at 3,000 sequential edits on ZsRE without backbone collapse~\cite{wang2024wise}, and MEMORYLLM, with a fixed memory
pool of about 1.07B parameters, maintains stable readout of newly injected context across 650,000 sequential context updates under random memory-slot replacement and exponential decay~\cite{wang2024memoryllm}. Because MEMORYLLM evaluates retention of recent context under continual slot turnover rather than simultaneous retention of all historical writes, and an update writes a context segment rather than a single fact edit, only coarse orders
of magnitude and protocol regimes are comparable. Even so, the gap of roughly one to two orders of
magnitude between dedicated modules and sequential backbone editing, and the gap between
batch and sequential writes, are consistent with a critical ratio shaped by
substrate and policy.

The functional form of the interference term also has direct support.
Forgetting during LoRA fine-tuning follows a shifted power law in the number
of fine-tuned parameters and update steps, and neither early stopping nor
fewer trainable parameters avoids it~\cite{kalajdzievski2024forgetting}. This
motivates a power-law form for $I_{s,\pi}$ as a starting point, to be tested
for dependence on the load ratio alone.

Write budget can partly substitute for capacity. GradMem fixes the memory at
eight vectors and increases the number of key--value pairs to be stored
(Figure~\ref{fig:pm-evidence}(c))~\cite{kuratov2026gradmem}. With one write
step, exact match falls from $100.0\%$ at 4 pairs to $32.6\%$ at 96 pairs;
with five write steps, it remains $99.1\%$ at 64 pairs and $88.4\%$ at 96
pairs. In Equation~\eqref{eq:pm-joint-scaling}, the write budget $U$ moves
the critical ratio to the right. This result comes from a synthetic task, and
existing studies vary either capacity or load but not both, so the third
prediction, collapse of load curves across capacities, remains untested.

\paragraph{Capacity demand and balanced growth.}
Analogous to Equation~\eqref{eq:ms-memory-demand}, maintaining a target
performance $p_0$ requires the load ratio to stay below its critical value,
which implies
\begin{equation}
 M_{\mathrm{req}}(E;p_0)\gtrsim
 \frac{I(E)}{\kappa\,\rho^\star_{s,\pi}}
 \propto \frac{E^{\nu}}{\rho^\star_{s,\pi}}.
 \label{eq:pm-memory-demand}
\end{equation}
The exponent $\nu$ has the same meaning as in
Equation~\eqref{eq:ms-memory-demand}, with $\nu\approx 1$ when experience consists
of independent facts and $\nu<1$ when it shares rules or skills. Parametric
memory compresses shared regularities into the same parameters as it writes,
so consolidation happens implicitly during training.
Equation~\eqref{eq:pm-memory-demand} leads to two differences from
non-parametric memory. First, when external capacity falls short, degradation
arises through retrieval competition or eviction while surviving records remain
intact; when parametric capacity falls short,
earlier content is directly overwritten in shared weights, which is harder to reverse. Second, $\rho^\star$
appears in the denominator, so improving the substrate or the write policy
can reduce the required capacity by orders of magnitude
(Figure~\ref{fig:pm-law}(c)), often more efficiently than adding parameters.
Within this constraint, the balanced growth path follows
Equation~\eqref{eq:ms-balanced-scaling}, with
$M/M_0\propto(E/E_0)^{\alpha_E/\alpha_M}$.

\paragraph{Write policy and lifecycle cost.}
Writing more often is not always better. TMEM varies how much context
accumulates before a parametric write is triggered, where F1 peaks at a 4K-token
threshold on LoCoMo (25.72) and at 12K on LongMemEval-S (41.24), and declines
at both smaller and larger thresholds
(Figure~\ref{fig:pm-conditions}(b--c))~\cite{ren2026tmem}. A plausible interpretation is that writing too often
produces fragmented updates and raises the effective load quickly, whereas writing
too rarely forces each update to compress too much context at once. The policy $\pi$ therefore
exhibits a task-dependent favorable range rather than shifting the critical ratio
monotonically.

Under the cost structure of Equation~\eqref{eq:ms-lifecycle-cost}, parametric
memory concentrates cost in $C_{\mathrm{build}}$ and $C_{\mathrm{maint}}$,
through gradient updates and batch consolidation, while
context-token readout overhead in $\overline C_{\mathrm{query}}$ is zero because readout happens within
the forward pass and consumes no additional prompt context. Its average cost
advantage therefore grows with the reuse count $N$. Under environmental
change, however, parametric memory cannot precisely delete or revise an
individual item, and each revision adds to the load, which raises maintenance
cost and directly affects adaptation (Section~\ref{sec:recall-not-learning}).
This suggests a natural division of labor, where rapidly changing experience is
retained in external memory, while stable, frequently reused experience is
consolidated into parameters, in line with the synergy discussed in
Section~\ref{sec:synergistic-cl}.

Parametric and non-parametric memory thus share resource-limited power-law
terms and an attainable frontier, but parametric memory differs in three
respects: its read budget is fixed by the architecture, its scaling relation
contains an interference term governed by the load ratio, and its capacity
demand is amplified by the critical ratio. Scaling parametric memory is
therefore not only a matter of increasing $M$; isolated writes, batch writes,
larger write budgets, and well-chosen write timing are needed to keep $\rho$
below $\rho^\star$.

%% file: sections/applications.tex
\section{Memory-Centric Continual Learning Applications Across Digital and Physical Domains}
\label{sec:applications}

When agents deploy over weeks or months, their operational context constantly shifts: user goals change, software interfaces update, and physical surroundings evolve. In these settings, continual learning serves two practical roles: it keeps long-running tasks on track, and it applies past experience to new problems. Memory provides the backbone for both. By storing past interactions, execution outcomes, and environmental states, memory gives agents a stable basis for future decisions.

\begin{figure}[!t]
    \centering
    \includegraphics[width=\linewidth]
        {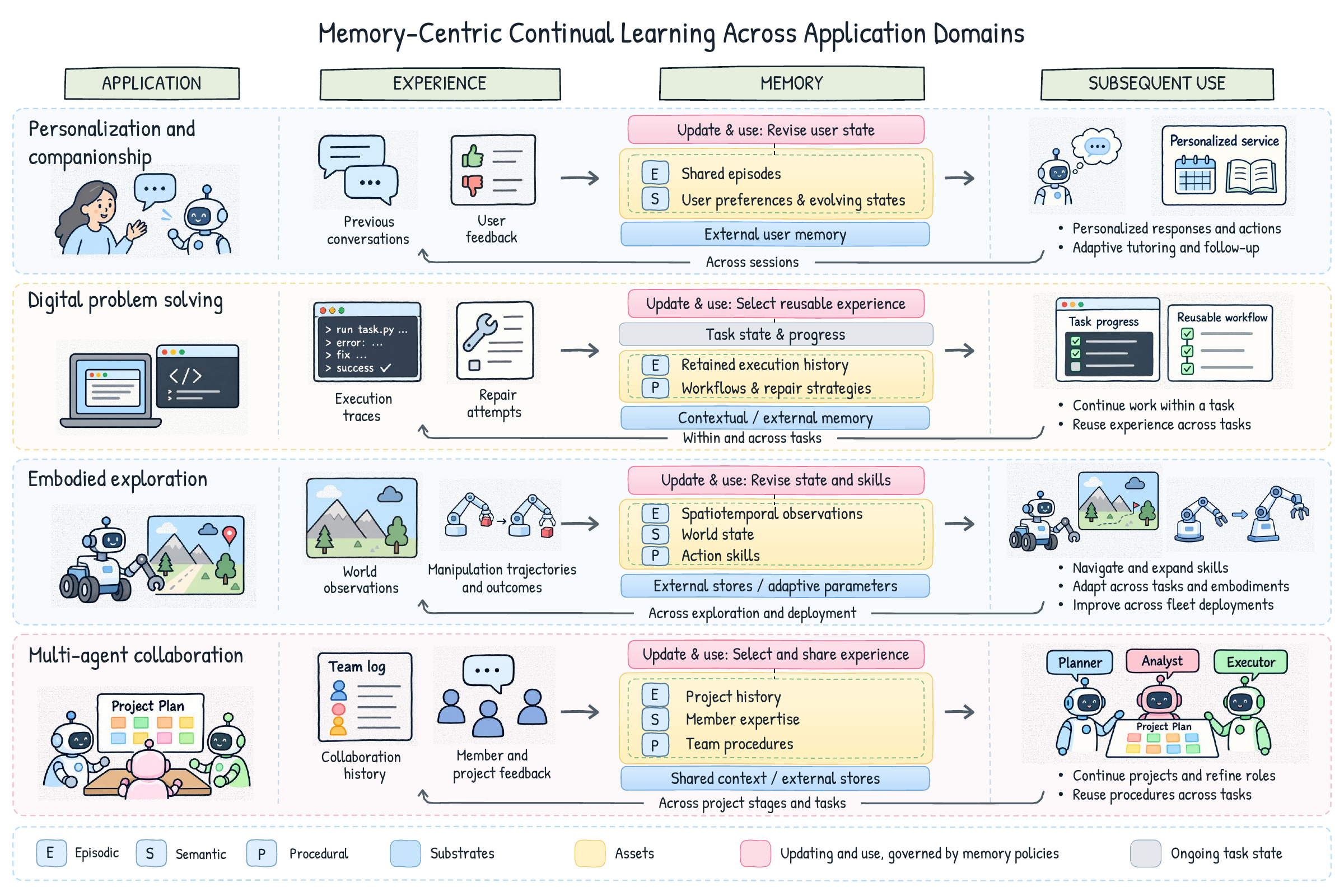}
    \caption{
        Memory-centric continual learning across application domains.
        Each row shows how accumulated experience updates memory to support
        personalized service, digital automation, embodied exploration, or
        multi-agent collaboration. Return arrows mark experience flowing back
        into memory across tasks and deployments.
    }
    \label{fig:applications-overview}
\end{figure}

These capabilities take four main forms across digital and physical domains (Figure~\ref{fig:applications-overview}):
(1) personal agents track evolving user preferences;
(2) digital agents turn execution logs into reusable workflows;
(3) embodied robots map environments and acquire physical skills; and
(4) multi-agent teams share project context and collaboration patterns.
Below, we examine how each setting turns accumulated experience into durable capability.

\subsection{Lifelong Personalization and Companion Agents}
\label{sec:application-personalization}

Personal agents must adapt as users change. Over time, schedules shift, learners build skills or form misconceptions, health symptoms evolve, and interpersonal bonds deepen. To remain useful, an agent must update its internal model of the user and use this state to guide dialogue, planning, and tool calls.

\noindent\textbf{Personal assistants.}\enspace
Assistants use cross-session logs to maintain conversational continuity and match individual style. Retrieving user profile items directly improves generation on tasks like email drafting and news summarization~\citep{salemi2024lamp}, while modeling complete interaction histories captures personal tone and habits~\citep{liu2025personaplug}. Crucially, assistants must track when facts expire. Recent systems update user facts across conversations~\citep{bae2022keep}, decay memory strength over time~\citep{zhong2024memorybank}, record user feedback alongside its valid context~\citep{yuan2025evolvingconditional}, and organize changing user states through causal links~\citep{ong2025theanine}. These mechanisms ensure that assistants preserve core preferences while promptly reflecting new changes.

\noindent\textbf{Tutoring agents.}\enspace
Tutoring agents track what a student knows and where they struggle. For example, ScaffoldLM stores pedagogical goals, step-level progress, and inferred misconceptions in an assessment memory, selecting the next hint based on the learner's latest response~\citep{li2026scaffoldlm}. Similarly, versioned student profiles track skill mastery, error patterns, and review schedules to calibrate hints and spaced practice~\citep{david2026intellicode}. Updating student profiles in real time allows agents to adjust lesson pacing, hint specificity, and problem difficulty on the fly.

\noindent\textbf{Health follow-up and emotional companionship.}\enspace
Memory directly underpins long-term care and companionship. In a ten-week study of 147 real-world users (1,252 calls), the CareCall system used memory to revisit past health complaints and life events. This continuity encouraged users to disclose more health details and build trust~\citep{jo2024carecall}. Studies on Replika users show that social bonding varies widely in depth and pace, meaning companion agents cannot rely on static facts alone~\citep{skjuve2022longitudinal,skjuve2023selfdisclosure}. Effective companionship requires referencing shared moments and conversational routines. For instance, recent agents infer implicit emotional needs and retrieve relevant shared memories to offer proactive comfort~\citep{fu2026enpmr}.

\noindent\textbf{From personalized generation to tool use.}\enspace
Personalization extends beyond text generation to planning and tool execution. Dual-reflection mechanisms build user memory banks that accurately route past preferences to current prompts~\citep{tan2025rmm}. In tool-augmented setups, benchmarks like Mem2ActBench test whether agents can translate past dates, locations, and habits into correct API parameters~\citep{shen2026mem2actbench}. Through memory, personalization becomes a continuous feedback loop: each interaction updates the user model, which in turn steers the agent's next action.

\subsection{Evolving Problem-Solving in Long-Horizon Digital Environments}
\label{sec:application-digital}

Digital tasks often span multiple web pages, applications, and execution steps. Web research requires tracking checked links and active filters; desktop tasks coordinate code editors, terminals, and spreadsheets; and software engineering requires navigating evolving codebases. In these settings, continual learning serves two needs: resuming interrupted workflows and transferring problem-solving strategies across tasks. Memory tracks progress to ensure task continuity, while abstracting successful runs into reusable skills and debugging routines.

\noindent\textbf{Web automation.}\enspace
Web navigation offers an ideal testbed for workflow reuse. For example, AWM distills reusable workflows from successful web trajectories and feeds them into future runs~\citep{wang2024awm}. Across 200 domains (such as shopping and travel), this approach raised success rates by 24.6\% on Mind2Web and 51.1\% on WebArena while cutting overall action steps. Online-learned routines---like filtering products or comparing flight routes---readily transfer to new websites with similar structures.

\noindent\textbf{Desktop and cross-application tasks.}\enspace
Desktop agents require both state tracking and skill reuse across applications. Systems like OS-Copilot store verified tool scripts in procedural memory~\citep{wu2024oscopilot}, while other architectures combine past execution traces with hierarchical planning to navigate unfamiliar interfaces~\citep{agashe2024agents}. Similarly, Trace2Skill distills raw trajectories into executable skill documents, demonstrating reuse across spreadsheet operations, vision-language tasks, and math reasoning~\citep{ni2026trace2skill}. For task continuity, OpenHands logs commands, file diffs, browser events, and user inputs into a unified event stream~\citep{wang2025openhands}. This complete history lets coding agents resume interrupted jobs without losing context. Together, these systems show that combining execution logs with procedural skills supports complex, multi-app workflows.

\noindent\textbf{Software engineering.}\enspace
Software engineering scales these demands to entire codebases. Coding agents must understand repository structures while logging inspected files, failed patches, and unit test outputs. SWEMem uses issue progress to selectively compress context, keeping context windows manageable during extended debugging sessions~\citep{gao2026swemem}. To transfer knowledge across bugs, SWE-Exp extracts debugging strategies from both successful and failed repairs to guide new GitHub issues~\citep{chen2025sweexp}. Thus, memory maintains local focus within a single bug fix while transferring repair strategies across the repository.

\noindent\textbf{Experience distillation and strategy reuse.}\enspace
Through trial and error, digital agents turn raw execution traces into reusable strategies. Recent frameworks extract high-level rules from repeated attempts, attach clear trigger conditions to guidelines, and index strategies derived from both successes and failures~\citep{zhao2024expel,fu2024autoguide,ouyang2025reasoningbank}. Terminal outputs, browser states, and test suites provide immediate feedback to test whether an extracted rule works. In short, memory gives digital agents two capabilities: it keeps long-horizon jobs from derailing, and it turns past mistakes into faster future execution.

\subsection{Continuous Exploration in Embodied and Open Worlds}
\label{sec:application-embodied}

Embodied agents must track a dynamic physical world while expanding their motor skills. Robots navigating homes must remember shifted furniture; robotic arms must adapt to unfamiliar tools; and open-world agents must set new goals as they explore. These tasks require two memory systems: a world-state memory that updates with visual observations, and a skill memory that stores reusable behaviors. Together, they allow agents to map their surroundings and master complex actions.

\noindent\textbf{Long-term navigation and scene understanding.}\enspace
Mobile robots rely on spatial memory to navigate and track dynamic scenes. For example, RememBR indexes video clips with spatial and temporal tags, allowing robots to answer queries about past object locations during navigation~\citep{anwar2024remembr}. Similarly, EmbodiedVideoAgent combines egocentric video, depth, and pose data into an object-centric memory, updating object states as interactions occur~\citep{fan2025embodiedvideoagent}. This structured memory boosts benchmark scores on Ego4D-VQ3D, OpenEQA, and EnvQA, and directly guides grasping, navigation, and placement. Historical observations thus become an editable 3D world model.

\noindent\textbf{Open-world exploration.}\enspace
In open worlds like Minecraft, agents must learn complex behaviors without human supervision. Voyager implements this by proposing new tasks, writing executable skill code, and saving verified skills into a library~\citep{wang2023voyager}. This skill bank helped Voyager discover 3.3$\times$ more items and unlock tech-tree milestones 15.3$\times$ faster than baselines. Crucially, the saved skill library lets agents solve unseen tasks from scratch in brand-new worlds. In text worlds like ScienceWorld, CLIN abstracts action outcomes into causal rules to guide future planning~\citep{majumder2023clin}. By saving executable code and causal patterns, exploration continually feeds subsequent skill acquisition.

\noindent\textbf{Cross-task and cross-embodiment adaptation.}\enspace
Manipulating physical objects requires sample-efficient learning. RoboCat addresses this by pooling experience across different simulated and real robot arms~\citep{bousmalis2023robocat}. The model learns a generalist policy, generates self-supervised data on new tasks, and feeds these runs back into training. Across this self-improving training loop, success rates on unseen tasks (after fine-tuning on 500 demonstrations) rose from 36\% with the initial model to 74\% with the final RoboCat model. Pooling data across diverse robot hardware shows that cross-embodiment memory drastically cuts the demonstrations needed for new tasks.

\noindent\textbf{Robot fleets.}\enspace
Robot fleets turn shared deployment into collective policy updates. In a 16-robot setup, autonomous trajectories and human teleoperation interventions are pooled to train a single generalist policy, which is then flashed back to the fleet~\citep{wang2026learningwhiledeploying}. Across eight manipulation tasks---including grocery restocking routines lasting 3--5 minutes---accumulated fleet data pushed success rates to 95\%. Field deployment thus acts as a continual training engine, linking individual robot trials to fleet-wide policy improvements.

\subsection{Collaborative Evolution via Shared Multi-Agent Memory}
\label{sec:application-multi-agent}

Multi-agent systems require shared memory to coordinate specialized roles. Software teams must log architectural decisions; data teams need to track data provenance; and scientific teams must record hypotheses, lab results, and reviews. Storing these records in a shared memory pool prevents redundant effort and lets agents reuse established workflows and team roles. Memory acts as the team's shared workspace, linking past findings to current tasks.

\noindent\textbf{Cross-task collaboration procedures.}\enspace
Shared procedural memory allows teams to reuse proven collaboration patterns. G-Memory organizes multi-agent interaction histories into a three-tier graph of dialogues, queries, and insights, letting agents retrieve both raw traces and abstract lessons~\citep{zhang2025gmemory}. Across five benchmarks, this structured memory improved embodied task success and question-answering accuracy. Similarly, LegoMem splits memory into orchestrator and worker tiers: orchestrator memory sharpens task decomposition, while worker memory improves tool-use accuracy~\citep{han2025legomem}. Reusing execution traces even helps teams of smaller LLMs match the performance of proprietary models.

\noindent\textbf{Member expertise and selective sharing.}\enspace
Teams also benefit from remembering which agent is good at what. LLMA-Mem builds a transactive memory that models member expertise, team layouts, and past coordination to optimize future role assignment~\citep{wu2026llmamem}. But sharing everything overwhelms agents. To prevent broadcast clutter, Fioresi et al. train an admission policy that only writes an intermediate step to global memory if it benefits downstream agents~\citep{fioresi2026learningtoshare}. On AssistantBench and GAIA, this selective sharing cut duplicated work without hurting task accuracy. In short, memory decides who should do the work and what information is worth broadcasting.

\noindent\textbf{Scientific collaboration.}\enspace
Scientific discovery teams use shared memory to iterate on complex research programs. In VirtualLab, specialized AI scientists hold multi-round meetings around a shared workspace to design nanobodies, guided by a human lead~\citep{swanson2025virtuallab}. Co-Scientist refines hypotheses across cycles of generation, reflection, and peer review, using persistent context memory to carry validation results forward~\citep{gottweis2026coscientist}. Similarly, Robin connects literature search, lab protocol design, and data analysis into an autonomous loop where wet-lab results update current hypotheses~\citep{ghareeb2026robin}. Shared memory keeps these long-running research pipelines coherent across iterative design cycles.

\noindent\textbf{Synthesis of shared-memory applications.}\enspace
Effective teamwork relies on three forms of continuity: project state (what the team is doing), member profiles (who is best equipped to do it), and shared procedures (how the work was done before). Project state keeps ongoing investigations on track; member profiles streamline role assignments; and reusable workflows eliminate redundant trial and error. Shared memory turns individual agent outputs into lasting organizational assets.

\noindent\textbf{Cross-domain synthesis.}\enspace
Together, these four domains highlight the practical scope of memory-centric continual learning. Applications now span multi-week health companions, cross-application web agents, repository-level bug fixing, open-world discovery, fleet-scale robotics, and automated science labs. In each case, memory links past experience to future action. By retaining and organizing execution traces, systems avoid repeating mistakes, adapt to changing users and environments, and continually turn raw interactions into durable capabilities.

%% file: sections/challenges.tex
\begin{figure}[!t]
    \centering
    \includegraphics[
        width=\linewidth
    ]{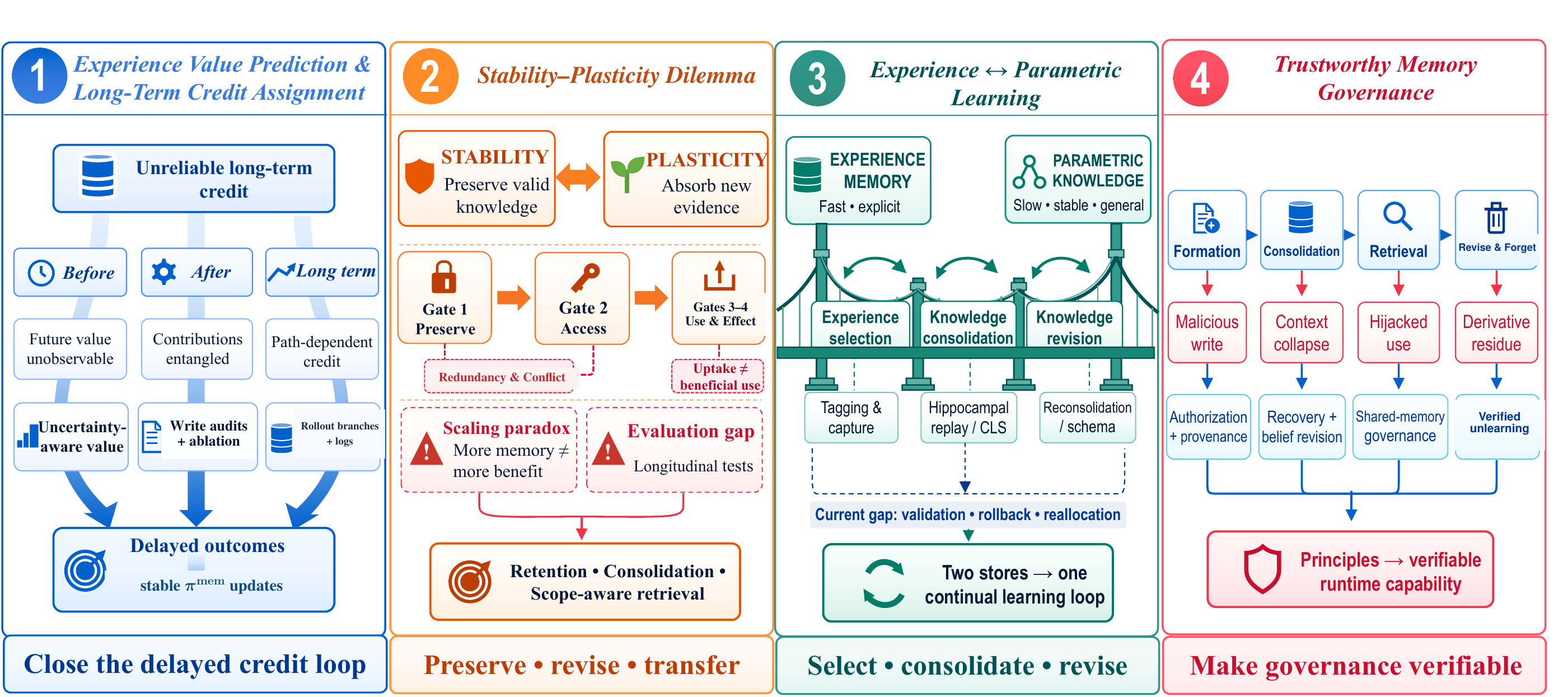}
    \caption{
        Four open challenges in memory-centric continual learning: 
        (1) credit assignment across prospective, retrospective, and long-term horizons; 
        (2) the stability--plasticity dilemma driven by memory interference; 
        (3) a bridge connecting episodic experience with parametric weights; and 
        (4) a lifecycle pipeline of deployment risks and governance rules. 
        Colors distinguish the four challenges, and the bottom bar highlights their core questions.
    }
    \label{fig:memory_challenges_landscape}
\end{figure}

\section{Open Challenges in Memory-Centric Continual Learning}
\label{sec:challenges}

\subsection{Predicting Experience Value and Long-Term Credit Assignment}
\label{sec:experience-value-credit}
Current agents still rely on handcrafted heuristics rather than learned policies $\pi^{\mathrm{mem}}$. The main obstacle is delayed credit assignment. When an agent records an event at step $t$, it must decide three things: whether to save it, how to format it, and what metadata to keep. But the true value of this write may not show up until step $t+100$ or later, after many rounds of summarization, pruning, and task shifts. As a result, final task success reflects a long chain of memory actions, not a single write. This delay blurs the credit signal. Agents can easily reflect using prompts, but they struggle to train policies that decide what to keep, update, or delete~\cite{bhargava2026memaudit,yan2026memory,agemem}. Feedback simply changes memory content ($\mathcal{M}_t \rightarrow \mathcal{M}_{t+1}$), while the policy itself stays fixed ($\pi_t^{\mathrm{mem}} = \pi_{t+1}^{\mathrm{mem}}$).

This challenge grows as memory moves from raw chat logs to reusable experience. Storing everything is impractical. Agents must weigh the future payoff of an experience against storage size, search latency, and the risk of confusion~\cite{luo2026storage,huang2026agentmemory,shu2026agentcl,zhang2026learning,wu2026forget}. Recent surveys map out how memory is written, managed, and read~\cite{du2026memory,wu2026memory,huang2026agentmemory,zhoubian2026memory,jiang2026anatomy}. Yet we still lack a principled way to balance these costs and benefits over time. Most existing methods rely on simple shortcuts: (1) rules based on recency, frequency, or text similarity~\cite{luo2026storage,wu2026memory}; (2) learning to call tools like store, search, or summarize~\cite{agemem,zhang2026memskill,zhang2025memoryaction,wang2025memalpha}; or (3) hard budgets on context size~\cite{zhang2026learning,wu2026forget,zhoubian2026memory}.

These shortcuts only look at the current prompt or immediate reward. They ignore delayed returns, distribution shifts, and interactions between co-retrieved memories. Saving bad data pollutes the store, triggers negative transfer, and hurts future tasks~\cite{shu2026agentcl,wei2025evomemory,hu2026memorycl,yang2026pathbench,yu2026forget,wu2026forget}. Three practical hurdles prevent delayed rewards from training $\pi^{\mathrm{mem}}$:
\begin{itemize}
    \item \textbf{Predicting value before writing.} At write time, the agent does not know future queries or tasks. Dropping an experience loses information permanently. Keeping bad data pollutes the store. The policy must forecast long-term utility using only local, present clues~\cite{bhargava2026memaudit,du2026memory,shu2026agentcl,wu2026forget}.
    \item \textbf{Tracing failure back through the pipeline.} Final errors stem from many sources: poor admission, lossy summarization, bad retrieval, or weak reasoning. When an agent fails, it is hard to isolate whether memory or reasoning caused the error~\cite{bhargava2026memaudit,jiang2026anatomy,du2026memory,ai2025memorybench,yan2026memory,zhang2026faultymemories}.
    \item \textbf{Path dependence across runs.} Every write or delete alters the memory state for future sessions. Different rollouts no longer share the same environment. This variance breaks standard reinforcement learning assumptions and destabilizes training~\cite{yan2026memory,yang2026pathbench,agemem,yu2026forget,zhang2026faultymemories}.
\end{itemize}

Solving these hurdles requires optimizing for long-term net value. For write-time prediction, agents can learn value models on task streams, using scaling curves to judge marginal returns~\cite{bhargava2026memaudit,shu2026agentcl,zhoubian2026memory,xiao2025densing}. To trace errors, systems can log write decisions, run ablation tests, and keep links from summaries back to raw source episodes~\cite{bhargava2026memaudit,du2026memory,jiang2026anatomy,shu2026agentcl}. To handle path dependence, agents can branch rollouts from identical memory checkpoints and use hindsight critics~\cite{yan2026memory,tan2026hindsight,agemem,zhang2026learning,wu2026forget}. The ultimate goal is closed-loop control: task feedback should update not just what the agent remembers, but how it manages memory.

\subsection{The Stability--Plasticity Dilemma in Dynamic Environments}
External memory gives agents a way to learn continually at the system level. Like classic continual learning, it aims to acquire new capabilities, preserve old knowledge, and transfer skills~\cite{luo2026storage,xiong2026learning}. As noted in Section~\ref{subsec:2.1}, parametric learning updates model weights, while agent memory updates external records at inference time. However, external stores do not remove the stability--plasticity dilemma; they simply relocate it~\cite{hu2026memorycl}.

Here, interference occurs during retrieval and prompt construction. A loose write policy (Section~\ref{subsec:formation}) boosts plasticity by saving new records quickly, but it also clutters storage with duplicate and conflicting entries. Under fixed retrieval budgets, this clutter crowds out useful memories. In the cascade of Section~\ref{sec:retention}, saving more records (Gate~1) hurts retrieval accuracy (Gate~2). Even when retrieved, outdated facts can lead the model to ignore correct updates or apply old rules to wrong tasks (Gate~3)~\cite{hu2026memorycl,liang2026learning,kim2026memory,zhang2026faultymemories}. Stability means keeping valid memories easy to find and use. Plasticity means updating old facts without breaking what already works.

This trade-off sharpens as memory expands. More interactions can help agents learn~\cite{ouyang2025reasoningbank}, and periodic cleaning keeps stores lean~\cite{yin2025experience}. Yet larger stores always raise the risk of retrieval noise. We should judge memory scaling by task gains under fixed compute budgets, not by raw storage volume.

Measuring these trade-offs requires long-term benchmarks. Recent benchmarks test specific skills, such as fact updates in LongMemEval~\cite{wu2025longmemeval} or task transfer in Evo-Memory and LifelongAgentBench~\cite{wei2025evomemory,zheng2025lifelongagentbench}. As outlined in Section~\ref{sec:evaluation}, evaluations should track retention, forward transfer, and negative interference across shifting task streams. Beyond final accuracy, tracking survival rates, retrieval ranks, and actual memory usage against frozen-memory baselines shows whether memory truly helps or quietly harms performance.

The core question is direct: \emph{How can agents keep memories accurate and relevant as experience grows?} Promising paths include write filters that prevent clutter, consolidation that checks raw evidence, and retrieval that matches the exact conditions where an experience applies~\cite{hu2026memorycl,zhang2026faultymemories,liang2026learning}.

\subsection{Synergistic Continual Learning through Experience Memory and Parametric Learning}
\label{sec:synergistic-cl}
Continual learning requires coordinating external experience with internal model weights. Today, agents treat them as two separate systems. External memory stores new facts quickly without touching model weights~\cite{shinn2023reflexion,chhikara2025mem0,xu2025amem}. Parametric fine-tuning bakes patterns into weights for broad generalization, but risks catastrophic forgetting~\cite{wu2024continual,shi2025continual}.

Recent work connects the two sides by extracting reasoning paths into external memory or fine-tuning lightweight adapters~\cite{gutierrez2025hipporag2,ouyang2025reasoningbank,zhao2024sapt,he2024seekr}. Yet information flows mostly one way: from raw episodes into weights or static stores. Systems cannot easily edit, roll back, or delete parametric knowledge once written. Human memory works differently. In biology, recalling a consolidated memory makes it flexible again, meaning the brain must actively restabilize it (a process called reconsolidation)~\cite{nader2000fear}. Neural networks lack this feature; once weights change, they stay fixed until the next gradient step.

The key missing piece is the \textbf{bridge between raw experience and parametric weights}. An agent should first test and summarize experiences in external storage. Only repeatedly verified patterns should enter model weights. Conversely, when existing weights fail on new evidence, the agent must pull that knowledge back into explicit text to inspect and fix it. This cycle follows Complementary Learning Systems (CLS) theory: fast learning in the hippocampus and slow integration in the cortex interact via replay to balance plasticity and stability~\cite{mcclelland1995why,kumaran2016learning,wilson1994reactivation}. Furthermore, the brain does not store every event; synaptic tagging-and-capture shows that novelty and reward flags decide which synapses undergo lasting change~\cite{frey1997synaptic}.

Future systems should build a two-way learning loop across three clear steps:
\begin{enumerate}
    \item \emph{Selection}: Filter raw traces by novelty, utility, and frequency---like synaptic tagging---before keeping them.
    \item \emph{Consolidation}: Replay, abstract, and bake verified memories into model weights via modular adapters.
    \item \emph{Revision}: Test weights against new facts, reopen outdated knowledge for correction (like reconsolidation), and use schemas to absorb new rules quickly~\cite{tse2007schemas}.
\end{enumerate}
Instead of treating memory and weights as separate silos, agents need clear algorithms that pass knowledge back and forth between them.

\subsection{Trustworthy Memory Governance in Lifelong Deployment}
While methods for writing and searching memory have matured, memory safety and control lag behind. Few studies address access control, rollback, or secure deletion, even though real-world deployment risks center on these exact areas~\cite{ding2026always}.

Without explicit governance, security risks spread across every stage of the memory lifecycle:
\begin{itemize}
    \item \textbf{Formation}: Attackers can inject malicious text through poisoned web pages or prompt injection. Without write checks, agents save these instructions as trusted facts, corrupting future actions (Section~\ref{subsec:formation})~\cite{chen2024agentpoison,dong2025minja,tian2026injecmem}.
    \item \textbf{Consolidation}: When agents summarize logs into general rules, they often drop timestamps and source links---a failure known as context collapse (Section~\ref{subsec:consolidation}). As a result, a single poisoned prompt can turn into a permanent, trusted behavioral rule.
    \item \textbf{Retrieval}: Corrupted memories can hijack tool calls and override user instructions during normal tasks~\cite{xu2026mcfa}, or leak private data across users in shared stores~\cite{yang2026crossuser}. This shows why retrieving a memory is different from using it safely (Section~\ref{subsec:retrieval_utilization}).
    \item \textbf{Forgetting}: Deleting a raw record does not remove its downstream traces, such as summaries, fine-tuned weights, or knowledge graphs (Section~\ref{subsec:revision_forgetting})~\cite{lin2026mnemonic}. Bad behavior can persist long after the original entry is gone~\cite{ding2026always}.
\end{itemize}

Memory control is hard for three reasons: (1) \emph{Expanding scope}: As agents gain more tools and autonomy, attack surfaces constantly shift~\cite{zhou2026externalization}. (2) \emph{Delayed impact}: The gap between a poisoned write and an attack can be weeks or months. Summarization hides original sources, making it hard to trace who wrote what~\cite{lin2026mnemonic}. (3) \emph{Direct trade-offs}: Strict privacy checks and audit logs slow down search and increase token costs~\cite{lam2026ssgm}.

Recent work introduces basic defenses: write permissions, provenance logs, user-isolated retrieval, and verified rollback~\cite{lin2026mnemonic}. Yet evaluation still falls short. Most benchmarks only score task success, giving developers little reason to build safe memory systems~\cite{ding2026always}. In addition, self-reflection remains hard to verify from the outside, and shared memory in multi-agent setups lacks clear access rules~\cite{wang2026unlearning,yang2026crossuser}.

To build reliable agents, governance must cover the full lifecycle:
\begin{itemize}
    \item \emph{Formation}: Verify sources and permissions before writing, rather than filtering data after problems occur~\cite{lin2026mnemonic}.
    \item \emph{Consolidation}: Borrow ideas from database recovery and belief revision to prevent agents from dropping critical source context during summarization~\cite{ding2026always}.
    \item \emph{Retrieval}: Sandbox tool execution and enforce role-based access for shared multi-agent stores~\cite{wang2026unlearning,yang2026crossuser}.
    \item \emph{Forgetting}: Use clear metrics---such as stress tests and recovery targets---to prove that deleted data and its side effects are truly gone~\cite{ding2026always}.
    \end{itemize}
Governance cannot remain an afterthought; it must become a testable, built-in guarantee of agent systems.

%% file: sections/conclusion.tex
\section{Conclusion}
\label{sec:conclusion}

In this paper, we reframe agent memory as an evolving learning
infrastructure for continual learning, rather than a passive storage or
retrieval component. We argue that experience accumulation cannot be equated
with static recall, but requires a closed-loop dynamic of retention,
adaptation, and transfer, explaining why current memory agents fail under
delay, interference, and environmental drift. To address this, we establish
a unified framework spanning physical substrates, reusable assets, and
lifecycle policies, coupled with a behavior-centric evaluation methodology
to isolate genuine experiential gains and a memory scaling law that relates
those gains to experience and memory resources. By adopting this memory-centric
lens, the field shifts its central question from whether an agent can
remember to whether it can continually learn, paving a more principled path
towards building resilient, self-improving intelligent systems.

%% file: sections/appendix.tex
\newpage
\phantomsection
\section*{Author List}
\markright{Author List}
\addcontentsline{toc}{section}{Author List}
\setcounter{section}{0}
\setcounter{subsection}{0}

\noindent
Dongfang Li$^{1,2}$, Zhuoen Chen$^{1}$, Bonian Jia$^{1,2}$, Zixuan Liu$^{1}$, Zhengxuan Lu$^{2,4}$, Can Deng$^{3}$, Yukun Shi$^{1}$, Yuhang Meng$^{1}$, Xin Zhang$^{1}$, Jiahe Huang$^{1}$, Fuhao Li$^{1}$, Junmai Wang$^{1}$, Shunyao Wen$^{1}$, Jingjun Sun$^{3}$, Jiaxu Tian$^{3}$, Jiangling Zhang$^{3}$, Zihan Wang$^{1}$, Liangyou Li$^{2}$, Yitong Yao$^{5}$, Shuangyong Song$^{5}$, Beilun Wang$^{4}$, Pengxu Wei$^{3}$, Hanlin Xu$^{6}$, Chang Liu$^{3}$, Yongxiang Li$^{5}$, Baotian Hu$^{1,2,*}$, Xiangyang Ji$^{3}$, Min Zhang$^{1,2}$

\medskip
\noindent
$^{1}$Harbin Institute of Technology (Shenzhen), Shenzhen, China\\
$^{2}$Shenzhen Loop Area Institute, Shenzhen, China\\
$^{3}$Tsinghua University, Beijing, China\\
$^{4}$Southeast University, Nanjing, China\\
$^{5}$Xingchen AGI Lab, China Telecom, Beijing, China\\
$^{6}$Huawei Technologies Co., Ltd., Shenzhen, China

\medskip
\noindent
$^{*}$Baotian Hu (\href{mailto:hubaotian@hit.edu.cn}{hubaotian@hit.edu.cn}) serves as the corresponding author for this paper.